%% file: main.tex
\documentclass{article} 
\usepackage{iclr2026_conference,times}

\input{math_commands.tex}

\usepackage{hyperref}
\usepackage{url}
\usepackage{booktabs}
\usepackage{longtable}
\usepackage{amsmath}
\usepackage{graphicx}

\input{numbers.tex}

\title{When the Reward Suite Is Leaky: \\ A Preregistered Causal Contrast
of Natural Verifier False Positives in RLVR}

\author{Chuyifei Zhang\\
Beijing Jiaotong University\\
\texttt{24222058@bjtu.edu.cn}}

\iclrfinalcopy
\begin{document}

\maketitle
\lhead{}
\renewcommand{\headrulewidth}{0pt}

\begin{abstract}
The test suites used as RLVR rewards for code have natural false
positives: per-task, persistent, asymmetric errors that accept the same
wrong programs every time they appear, unlike the symmetric or resampled
noise assumed by existing noise-robustness analyses.
We run a preregistered two-arm causal contrast on a deployed suite: GRPO
on identical MBPP tasks, seeds, and compute
(Qwen2.5-Coder-1.5B-Instruct, \nseeds\ seeds $\times$ \nsteps\ steps),
rewarded by the original MBPP tests (leaky) versus the MBPP+ extra tests
(hardened; extra-tests-only, \S\ref{sec:arms}). Two further families
replicate the design under a preregistration
frozen before their data existed: deepseek-coder-1.3b-instruct,
registered as the decision arm, and Llama-3.2-1B-Instruct. Claims below
are tagged \regC\ (confirmatory: preregistered) or \regE\ (exploratory).
\regC\ The average held-out effect (scored by the extra-test suite
throughout) is bounded: non-inferior under a
preregistered \Xmargin-pt margin (gap \pgapmean\,pt, one-sided 95\% upper
bound \pgapupper\,pt), with the upper bound on the same side of the
margin under all six estimators in all three families.
\regC\ Rewarded false-positive mass tracks a cheap static leakiness audit
computed before training (Spearman \rhotrackraw\ raw, \rhotrackpartial\
under difficulty control), and the registered train-side test puts the
leak-stratum FP share \ptrainshare\,pt above clean tasks (CI90 \ptrainci;
the FP definition is classifier-free, so these numbers are invariant
under all later audit revisions).
\regE\ Auditing every rewarded FP under signed, human-adjudicated rules
finds a large residual of verified genuinely wrong code: \vwsharevfive\%
record-weighted (task-cluster bootstrap-95 \vwclusterci; estimator
variants within two points). The reward paid for real bugs, not merely
suite artifacts. Both replication families reproduce a large residual
share --- \famBvwsharevfive\% and \famCvwsharevfive\% --- but the shares
replicate ``large,'' not one number.
\regE\ On statically leaky held-out tasks the leaky arm improves less
(family A). This is an association only, its specification-search
correction prices one axis of freedom (a lower bound), and it does not
replicate cleanly across families: one reversed-direction candidate, one
underpowered positive.
\regE\ Mechanism evidence is consistent with selection of pre-existing
error modes rather than learned exploitation. FP incidence sits at its
long-run share from step 0 and does not grow within our horizon (flat in
two families, declining in the third); the largest channel shows
\distinctlargest\ distinct wrong programs and no hacking signatures; and
untrained base models already produce the same wrong outputs under the
leaky filter. The \rewardgap-pt reward inflation shows no held-out
counterpart within the bound reported above, though parameter-level
sharpening below behavioral resolution is not excluded. In an exploratory
follow-up outside these confirmatory claims, we turn the same instrument on
the frontier judges themselves: on their own false positives they
self-assess only weakly, a deconfounded same-author test is unresolved, and
even the highest-scoring reader we probe stays far below its score on a weaker
policy's errors --- two subjects on MBPP, licensing nothing about frontier
models in general (\S\ref{sec:meta}).
The practical boundary we map: a cheap static audit locates where FP mass
will sit before training --- exposure, not final damage. Hardening the
reward --- swapping the leaky tests for the far larger extra-test suite
--- removes the measurement inflation, though at this scale it buys
little capability.
\end{abstract}

\section{Introduction}
\label{sec:intro}

Reinforcement learning from verifiable rewards (RLVR) trains code models
against test suites and treats a pass as ground truth
\citep{lambert2024tulu,guo2025deepseekr1}. Deployed suites do not earn that
trust: a recent audit of two public code-RL benchmarks found that
25--28.5\% of audited tasks accept at least one wrong solution
\citep{rajan2026auditing}. That audit, however, measures the suites, not
what happens when models train against them, and it names the missing
experiment: ``a direct causal test (training on fixed-versus-untouched
broken tasks at fixed compute) is the natural next step.'' This paper runs
that test.

The verifier errors we study are not label noise. When a deployed suite
accepts a wrong solution, it accepts it every time---the same wrong program
passes the same weak tests on every rollout. These natural false positives
(FPs) are therefore per-task and persistent, and the error runs in one
direction only: wrong code gets rewarded. Noise-robustness studies model
verifier error differently, as random flips of the pass/fail signal that
are redrawn on every attempt. Because such noise averages out over
training, RLVR tolerates up to 15\% of it with little loss
\citep{plesner2026imperfect}. A persistent FP does not average out; it is a
standing reward for a specific wrong program on a specific task. Whether
trained policies collect that reward, and at what cost, is what we measure.

\begin{figure}[t]
\centering
\includegraphics[width=\textwidth]{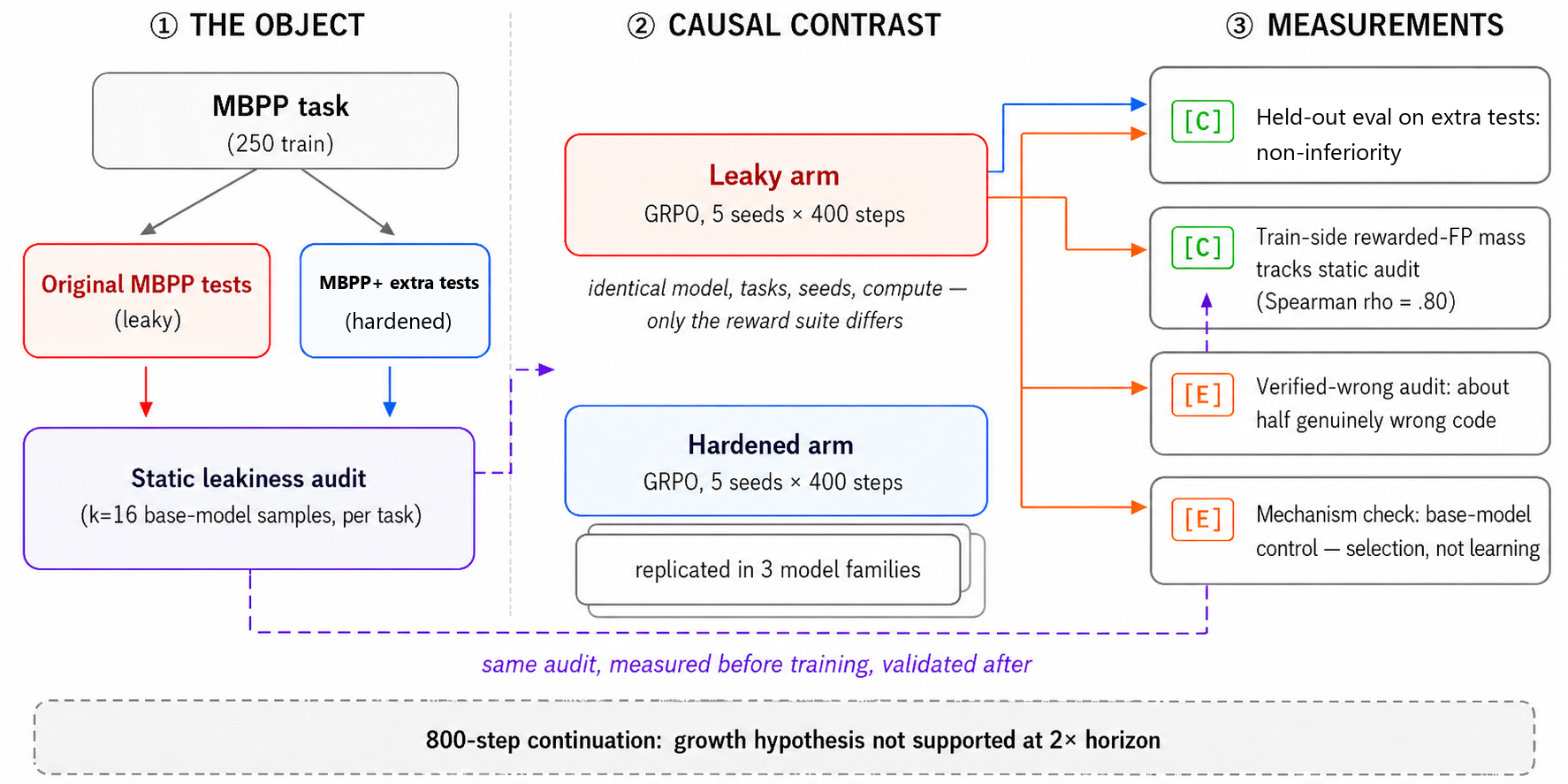}
\caption{\textbf{Overview of the design.} \emph{Left:} every MBPP task
ships two suites --- the deployed, leaky original tests and the hardened
arm's extra tests (the MBPP+ ``plus'' tests) --- and a static audit scores per-task leakiness from
$k{=}16$ base-model samples before any training (\S\ref{sec:arms}).
\emph{Center:} two GRPO arms differ in exactly one component, the suite
that pays the reward, holding model, tasks, \nseeds\ seeds, and compute
fixed; the whole design repeats in three model families
(\S\ref{sec:xfam}). \emph{Right:} held-out pass rate is always scored by
the extra-test suite (MBPP+ ``plus'' tests, extra-only; non-inferiority, \regC, \S\ref{sec:avg}); rewarded-FP
mass measured during leaky-arm training tracks the static audit
(Spearman $\rho=\rhotrackraw$, \regC, \S\ref{sec:track}) --- the bold
dashed arrow marks the audit's double duty, scoring leakiness before
training and validated against what training pays for after, while the
thin dashed arrow marks the same audit stratifying the two-arm
comparison (\S\ref{sec:strat}); an exhaustive audit of every rewarded
false positive in family A finds about half genuinely wrong code (\regE,
\S\ref{sec:vw}); and a base-model control finds the same wrong outputs
before training, consistent with selection of pre-existing error modes
(\regE, \S\ref{sec:mech}). A preregistered continuation finds no growth
at twice the horizon (\S\ref{sec:cont800}). \regC\ marks confirmatory
(preregistered) results, \regE\ exploratory (post-hoc, labeled) ones.}
\label{fig:pipeline}
\end{figure}

We measure it with a preregistered two-arm experiment on a suite people
actually train on. Both arms train the same model
(Qwen2.5-Coder-1.5B-Instruct; \citealp{hui2024qwencoder}) with GRPO
\citep{shao2024deepseekmath} on the same \ntrain\ MBPP tasks
\citep{austin2021program}, with the same seeds, hyperparameters, and
compute; only the reward differs (Figure~\ref{fig:pipeline}). The
\emph{leaky arm} is scored by the
original MBPP test suite, while the \emph{hardened arm} is scored by the
extra tests that the matching MBPP+ project \citep{liu2023evalplus}
generates for the same tasks --- roughly a hundred automatically
generated tests per task against MBPP's three --- run \emph{instead of}
the originals, not on top of them.\footnote{``Hardened'' throughout this
paper means \emph{extra-tests-only}: the reward, and all held-out
scoring, runs the MBPP+ ``plus'' tests alone, as preregistered --- not
the official EvalPlus metric, which requires passing base \emph{and}
extra tests together. Nor does either test set contain the other: on
\disjointmbpp\ of \mbpptotal\ MBPP tasks the base and extra sets are
fully disjoint. \S\ref{sec:arms} defines the arms; the
suite-operationalization paragraph in \S\ref{sec:limitations} quantifies
the consequences.} Before training, we map where the
leaky suite fails with a cheap static audit: we sample solutions from the
base model and flag those that pass MBPP but fail MBPP+. We then train
\nseeds\ seeds for \nsteps\ steps per arm, and we evaluate both arms on
held-out MBPP and HumanEval+ tasks, always scored by the same extra-test
suite.
Finally, two more model families---deepseek-coder-1.3b-instruct
\citep{guo2024deepseekcoder} and Llama-3.2-1B-Instruct
\citep{metaai2024llama32}---repeat the whole design under a preregistration
frozen before their data existed.

Every claim in this paper carries one of three labels. \regC\ marks
confirmatory results, preregistered before the relevant data existed.
\regE\ marks exploratory results, computed after seeing the data and
labeled as such wherever they appear. \regI\ marks instrument-validity
checks, which we publish no matter how they turn out. The
confirmatory--exploratory distinction follows preregistration practice
\citep{vanmiltenburg2021preregistering}; the labels simply keep it visible
next to each claim. We put the split up front because our strongest-looking
numbers do not all carry the same evidential weight, and the reader should
never have to guess which is which. Everything below holds at our scale and
horizon---\nsteps\ GRPO steps, 1--1.5B models---and we claim nothing past
that. A separate, exploratory meta-audit (\S\ref{sec:meta}) turns the same
instrument on the frontier \emph{judges} that grade this kind of code; it
sits outside every confirmatory claim here and generalizes to nothing beyond
its two subjects.

The contribution is the measurement layer itself: a preregistered
causal contrast that isolates what a deployed suite's natural false
positives do to training, with instruments that separate where reward
leaks, what it pays for, and whether training climbs it. Within that
scope, we find:
\begin{itemize}
\item \textbf{The average held-out effect is bounded \regC.} At step
\nsteps, the hardened arm ends \pgapmean\,pt ahead of the leaky arm on
held-out pass rate (CI90 \pgapci), and the one-sided 95\% upper bound on
that gap---\pgapupper\,pt---sits inside the preregistered margin of
\Xmargin\,pt under all six estimator variants. The leaky arm's rewarded-FP
mass also stays flat across training, and both arms improve over their base
model (\S\ref{sec:avg}). Noise-robustness results
\citep{plesner2026imperfect} do not predict this, because persistent FPs do
not average out. The bound is also one-way: it says what did not happen in
\nsteps\ steps at this scale, not what cannot happen
(\S\ref{sec:limitations}).
\item \textbf{A cheap static audit predicts where the reward leaks \regC.}
Rewarded-FP mass per task tracks the audit's leakiness score, with Spearman
\rhotrackraw\ raw and \rhotrackpartial\ after controlling for task
difficulty. On the training tasks the audit flags as leaky, the leaky arm's
FP share runs \ptrainshare\,pt above clean tasks (CI90 \ptrainci;
registered test; \S\ref{sec:track}, \S\ref{sec:strat}).
\item \textbf{About half of the leaked reward pays for genuinely wrong code
\regE.} We audited all \vwnvfive\ rewarded FPs in family A under written,
human-adjudicated rules. Of these, \vwsharevfive\% (record-weighted) are
verified-wrong---real bugs, not suite artifacts---with a task-cluster
bootstrap-95 interval of \vwclusterci, and every estimator variant lands
within two points of that figure (\S\ref{sec:vw}). Both replication
families reproduce a large share.
\item \textbf{Mechanism: consistent with selection, not learning \regE.}
Three observations point the same way. The leaky arm's FP rate is already
at its long-run level within the first training steps, and it does not grow
over the horizon we measure. The wrong outputs the suite pays for already
show up in samples from the untrained base models. And the extra reward the
leak buys (\rewardgap\,pt) comes with no measured change in held-out
ability. Together, these are consistent with selection of pre-existing
error modes, not learned exploitation (\S\ref{sec:mech}). Whether anything
grows at larger scale or longer horizons is a falsifiable continuation
hypothesis, and our data does not answer it (\S\ref{sec:limitations}).
\item \textbf{Statically leaky held-out tasks improve less, in one family
\regE.} In family A, the held-out tasks our audit flags as leaky gain less
under the leaky arm than under the hardened arm. That result comes with
three qualifiers. It is an association, not a causal estimate. The
correction we ran for analysis freedom covers one axis only, so it is a
lower bound. And it does not replicate cleanly across families: one shows a
reversed-direction candidate signal, the other an underpowered positive
(\S\ref{sec:strat}).
\item \textbf{Two more families replicate the core measurements \regC.} All
three preregistered replication criteria pass on the family we preregistered
as decisive (deepseek-coder-1.3b): a large verified-wrong share, leakiness
tracking, and the train-side stratification gap. The third family
(Llama-3.2-1B) shows the same tracking and stratification
(\S\ref{sec:xfam}). We release the audit tooling, the adjudication record,
and the full preregistration chain.
\end{itemize}

\section{Related Work}
\label{sec:related}

\textbf{Verifier noise in RLVR.} Robustness studies model verifier error
as random corruption and find RLVR forgiving of it:
\citet{plesner2026imperfect} inject symmetric, per-epoch-resampled flips
plus model-judge noise and see little damage up to 15\%;
\citet{rad2026rateorfate} supply the aggregate theory, a phase transition
in the verifier's Youden index (\S\ref{sec:discussion}). Both treatments
average over tasks, and \citet{plesner2026imperfect} say plainly what
their noise is not: ``real verifiers have asymmetric errors and
persistent biases.'' Those persistent, per-task errors are our object.
\citet{egashira2026delay} come closest on the error model: in controlled
arithmetic experiments, systematic false positives --- unlike random ones
--- drive outcomes from sub-optimal plateaus to collapse, and the damage
follows the error pattern, not the aggregate rate. Their errors are
injected; ours ship with a deployed suite. Corrections for noisy
verifiers are orthogonal to our measurement question
\citep{cai2025imperfect}, as are demonstrations that headline gains can
be illusory without any verifier error: random rewards buy large
MATH-500 gains through a GRPO clipping bias \citep{shao2025spurious},
and budget mismatch, attempt inflation, and contamination overstate
reported gains \citep{wu2025position}. The resolution below the
aggregate --- which tasks leak, what the leak pays for, whether the
policy climbs it --- is where this paper lives.

\textbf{Known suite errors in code benchmarks.} That deployed test
suites err is on record at the single-task level: the EvalPlus issue
tracker documents wrong reference solutions and wrong tests on
individual MBPP+ tasks.\footnote{Issues \#210 (\texttt{Mbpp/459}, 2024)
and \#301 (\texttt{Mbpp/102}, 2026) at
\url{https://github.com/evalplus/evalplus/issues}.}
\citet{stroebl2024inference} price such errors at inference time: false
positives put a ceiling on resample-until-pass scaling that no compute
budget lifts, and the MBPP and HumanEval unit tests they study have
limited coverage. On the repair side, \citet{ma2025rethinking} quantify
test-suite thoroughness, showing that sparse, homogeneous tests inflate
measured performance and distort RLVR reward estimation, and generate
stronger tests; CodeHacker \citep{shi2026codehacker} generates
adversarial tests that filter previously accepted wrong solutions in
competitive programming. Contemporaneous RLVR pipelines, meanwhile,
still adopt the raw fraction-of-tests-passed signal as their reward on
MBPP without examining suite adequacy \citep{kim2026compete}. We
neither repair a suite nor report single tasks: we measure, per task and before training, where a deployed
suite's false positives sit, then verify against training what the leak
pays for.

\textbf{Audit-side measurement.} \citet{rajan2026auditing} audits the
acceptance rate of deployed code-RL suites (SWE-bench Verified and
R2E-Gym; 25--28.5\% of audited tasks
accept a wrong patch) and shows observationally, across 134 public model
submissions, that reported pass rates are inflated on statically hackable
tasks. That audit measures suites, not training, and leaves the causal
question open by name (\S\ref{sec:intro}). We run the experiment that
closes the gap, and add the piece a dataset curator needs before
training: a per-task prediction of where reward will leak, validated
against training-time FP mass in three model families.

\textbf{Engineered exploits versus natural FPs.} Existing coding-RL
hacking testbeds study active exploits that leave a structural artifact
in the emitted solution: rewriting the evaluator \citep{wu2026rebounds},
hardcoding known test cases \citep{taufeeque2026obfuscation}, planted
exploit surfaces such as always-equal \texttt{\_\_eq\_\_} classes,
\texttt{sys.exit}, and \texttt{conftest} patches \citep{beigi2026prime},
or hacking seeded through contaminated distillation data
\citep{khalifa2026countdown}. Our setting has no such structural exploit
artifact: leaky unit tests are genuinely passed by wrong solutions that
read as honest attempts, wrong only in their correctness. The nearest of these testbeds mentions the natural case only
in passing, as something that ``rarely occurs''
\citep{taufeeque2026obfuscation}; we make our neighbors' footnote the
object of study, and find that close to half of the reward it leaks pays
for genuinely wrong code.

\textbf{Hardening and detection tooling.} A concurrent tool-building
line attacks the same weakness from the repair side: fuzzing RLVR
verifiers to surface false-positive regions before training
\citep{ray2026fuzzing}, adversarial hacker--fixer loops that patch
agent-benchmark verifiers until known exploits stop passing
\citep{zhong2026hardening}, isomorphic-invariance tests that catch the
shortcut solutions extensional verifiers admit \citep{helff2026gaming},
and benchmarks scoring how well models detect reward-hack trajectories
\citep{deshpande2026trace}. We build no such defense; we measure what
the undefended suite does to training, which is the baseline any repair
must beat.

\textbf{Instruments, and the selection--manufacturing gap.} Prior work
instruments the capability side of reward hacking: chain-of-thought
monitors, elicitation probes, and difference-of-means activation vectors
that rise under proxy RL and are ablatable \citep{beigi2026prime};
deception probes and representation drift
\citep{taufeeque2026obfuscation}; activation-level monitoring during
generation \citep{wilhelm2026monitoring}. We instrument the sampled
behavior instead: the error distribution of emitted solutions, measured
against a base-policy control. The control matters because rising probe
scores under RL do not by themselves distinguish \emph{manufacturing} a
propensity from \emph{selecting} one already present in the pretraining
prior --- pure selection also inflates on-policy readings.
Representation-level evidence in fact leans toward pre-existence: concept
directions extracted from the base model remain geometrically unchanged
(cosine $\geq 0.99$) across RL \citep{wu2026rebounds}, and code
correctness is decodable from the internal states of models that have
never seen RL \citep{bui2025openia}, with correctness self-assessment the
first component to move under proxy RL \citep{beigi2026prime}. But
direction stability speaks to the probe axis, not to whether the base
policy already emits the exploiting behaviors, and none of these lines
samples the initialization on the same tasks and verifier. Our base-model
control supplies that discriminant at the behavioral level
(\S\ref{sec:mech}). The selection reading also has concurrent
support from a different reward channel: \citet{zhouc2026convincing}
find that the inflation produced by self-play against a reference-free
LLM judge replicates without any parameter update under best-of-$N$
selection, in code as in competition math. Their reward is a learned
judge's plausibility preference, ours a fixed test suite with natural
coverage holes; the two substrates independently return the same
selection-over-manufacturing reading.

\textbf{When judging is not easier than generating.} The meta-audit of
\S\ref{sec:meta} sits in a growing literature on verification limits.
\citet{dorner2025limits} prove that a judge no more accurate than the
model it evaluates cannot cut ground-truth labeling needs by more than
half, whatever the debiasing; \citet{zhou2025variation} find that
stronger generators produce errors that are harder for verifiers to
detect, across math, knowledge, and language reasoning;
\citet{bandyopadhyay2026judge} find self-evaluation trailing generation
on three of four in-context QA benchmarks; in self-evolving agent
pipelines, a judge's false-pass bias can silently disable a downstream
curation mechanism while aggregate metrics stay flat
\citep{zhang2026blindcurator}. On code, EvilGenie reports
LLM judges detecting unambiguous reward hacking well
\citep{gabor2025evilgenie}. Our exploratory meta-audit adds the
code-reward false-positive case, with self- and cross-judge grids over
two frontier subjects; it is descriptive, and two subjects on one
benchmark license nothing about frontier models in general.

\textbf{Authorship signals in LLM judges.} Whether a judge treats its
own output differently is usually studied without telling the judge
whose output it reads. \citet{mahbub2025obfuscation} perturb candidate
outputs to obscure authorship and find self-preference drops, then
recovers as stylistic differences are neutralized further;
\citet{khullar2026selfattribution} find monitors more lenient on actions
implicitly framed as their own by turn structure, while an explicit
self-attribution statement alone did not induce the bias;
\citet{guey2026selfpreference} find no detectable self-preference when
models judge verifier-validated revisions of their own drafts. Our
attribution study (\S\ref{sec:meta}) holds the code fixed and
manipulates an explicit ownership claim. We detect no self-protective
shift in this pipeline; the one Holm-surviving effect points in the
harsher-when-truly-own direction, consistent with truth-sensitivity of
the response through any channel and, inseparably, with a
claim-content$\,\times\,$pool-property interaction that is not
truth-mediated. The designs differ --- implicit framing versus explicit
claim text, safety monitoring versus correctness judging --- so the
directions are not directly comparable.

\textbf{Growth, scale, and stakes.} Where exploitation is measured over
training it often grows: rubric exploitation intensifies across training
\citep{mahmoud2026rubric}, and the verification gap widens with task
length --- 28 percentage points per tenfold increase in code size in
long-horizon agent settings \citep{zhao2026specbench}. Our
selection-not-learning result sits at the other end of both axes (short
tasks, \nsteps\ steps, 1--1.5B), and we claim nothing beyond that scope
(\S\ref{sec:limitations}); the position that no fixed reward survives
capability growth \citep{wang2026verification} is the scale-axis version
of our continuation hypothesis. The stakes of the quiet case are on
record: a model that does learn to hack can generalize to broad
misalignment \citep{macdiarmid2025misalignment}, and the most dangerous
variant leaves no pass-rate footprint --- in \citet{beigi2026prime}'s
gold-reward branch the hack rate falls to 3\% while the internalized
capability persists. Our natural FPs lack the cheating semantics that
mediate the misalignment result --- a mechanism our design does not test;
the lesson we draw is narrower: a clean aggregate is not evidence of a
clean reward. Concurrent mitigation work is orthogonal
--- \citet{damani2026right} add a demonstration-based discriminator to
the reward --- and our contribution is the measurement layer any such
defense would be evaluated against.

\section{Setup}
\label{sec:setup}

\subsection{Arms, suites, and training}
\label{sec:arms}

Our unit of comparison is a pair of training arms that differ in exactly
one component: the test suite that pays the reward. Every MBPP task ships
with a small set of assert-style tests (\basetestsavg\ per task on
average), and the EvalPlus project extends the same tasks with a much
larger, automatically generated set, MBPP+ \citep{liu2023evalplus}
(\extratestsavg\ per task); we pin evalplus 0.3.1 throughout. The
\emph{leaky arm} rewards a rollout if it passes the task's original MBPP
tests (\texttt{base\_input}). The \emph{hardened arm} rewards it if it
passes the MBPP+ \emph{extra} tests (\texttt{plus\_input}) --- the extra
tests only, as preregistered. This is not the official EvalPlus metric,
which requires passing base and extra together; nor does either suite
contain the other --- on \disjointmbpp\ of \mbpptotal\ tasks the base and
extra sets are fully disjoint. Prompts, tasks, seeds, decoding, optimizer
settings, and step counts are identical across arms, so a difference in
outcomes is attributable to which suite pays the reward.

Training uses GRPO \citep{shao2024deepseekmath} with group size 8, sampling
temperature 1.0, a 640-token completion cap, and no KL anchor ($\beta=0$,
no reference model), for \nsteps\ steps on \ntrain\ MBPP training tasks,
\nseeds\ seeds per arm, with checkpoints every 100 steps. Family A, the
family every audit in this paper goes deepest on, trains
Qwen2.5-Coder-1.5B-Instruct \citep{hui2024qwencoder}. Two replication
families repeat the whole design under a preregistration frozen before any
of their data existed (\S\ref{sec:prereg}): family B is
deepseek-coder-1.3b-instruct \citep{guo2024deepseekcoder}, preregistered as
the decision family for the replication criteria, and family C is
Llama-3.2-1B-Instruct \citep{metaai2024llama32}.

Before any training, a static audit maps where the leaky suite fails. For
each task we draw 16 samples from the family's own base model and score
every sample against both suites; a task's \emph{leakiness} is the fraction
of its MBPP-passing samples that fail MBPP+ (undefined when nothing
passes). The audit executes code but involves no training, and it is
cheap: one sampling pass per family.

\subsection{Splits and evaluation}
\label{sec:evalproto}

We froze a problem-level split before the first run: \ntrain\ MBPP training
tasks and \nevalmbpp\ held-out MBPP evaluation tasks, disjoint by task,
plus \nevalhe\ HumanEval+ tasks as an out-of-distribution set (split RNG
seed \splitseedv). Held-out performance is always scored by the
extra-test suite (the MBPP+ ``plus'' tests, extra-tests-only), whichever
arm produced the model. At each checkpoint we draw
$k=\ktemp$ samples per task at temperature 0.8 and report mean pass rate
over the \nevalpaired\ paired tasks; the seed is the inference unit
throughout (\S\ref{sec:discipline}).

Two train-side quantities recur. A \emph{rewarded FP} is a leaky-arm
rollout that passes the MBPP tests, and therefore collects reward, but
fails MBPP+. \emph{FP mass} is the fraction of rewarded rollouts that are
FPs. Both are mechanical: they read the two suite verdicts and nothing
else, so they do not depend on the audit labels of \S\ref{sec:vwspec} and
are invariant under every later re-classification.

\subsection{Preregistration, including its failures}
\label{sec:prereg}

The registration history publishes with this paper, failures included.
Version 1 froze the split, the non-inferiority margin, the step and seed
counts, and the outcome directions before any grid run. Its launch gate, a
blind hand-audit of 24 sampled FPs, was voided when \gatebvoid\ verdicts
turned out to hinge on properties of the test suites rather than on scorer
errors; the accompanying diagnosis established that the scorer itself
agreed with the official evalplus harness on all 24 audited samples
(\scorerdiv\ divergences), an instrument check we report as
instrument-validity \regI. The one-shot replacement gate then failed as
well: \gatebprime\ of its verified-wrong samples were overturned on
evidence. That failure was predictable from the signed document's own
numbers, and we record it as a design error rather than bad luck. No
outcome data had been observed at either point; both failures concern the
labeling instrument (which plus-fails count as genuinely wrong code), not
the outcomes.

Version 2 responded in two ways. It replaced sampling gates with an
exhaustive audit of the full FP pool (\S\ref{sec:vwspec}), and it added a
gate-power rule: no future gate may be signed without a published
calculation showing at least 80\% pass probability when the instrument is
in fact correct. Version 2 also pre-wrote the claim ladder for the audit's
outcome: the verified-wrong share selects one of three pre-committed
sentences, and since every added artifact predicate can only shrink the
share, iterating on the audit spec moves the result against our own
exploitation claim. The audit is self-penalizing by construction. Version 3
registered families B and C, the replication criteria, and the family-B
keying before any replication data existed.

\subsection{The verified-wrong audit}
\label{sec:vwspec}

The mechanism section needs a label that no test suite provides: of the
rollouts the leaky reward paid, which contain genuinely wrong programs, as
opposed to correct-enough programs that a defective extra test rejects?
The train-side FP pool is finite (\vwnvfive\ records in family A), so we
audit all of it rather than sample. A written predicate specification
classifies every record as verified-wrong or as one of four non-model
causes: plus-suite false negatives (the extra tests themselves reject
correct code), contract-invalid test inputs, canonical-convention
mismatches, and resource-limit artifacts. Predicates are per-task and
evidence-backed, each signed by the human adjudicator, and the analysis
scripts refuse to emit any verified-wrong aggregate from grid data unless
the signed marker file exists. The specification went through four signed
generations; the family-A headline share fell \vwshare\% $\to$
\vwsharevthree\% $\to$ \vwsharevfive\% along the way, and
\S\ref{sec:vw} reports that trail, not just its endpoint. One lesson from
the audit is itself a method: ``both arms score zero on the plus tests''
is necessary but not sufficient evidence that the suite, rather than the
model, is broken --- five seeds sharing one wrong convention produce the
same signature.

\subsection{Honesty notes on the instruments}
\label{sec:honesty}

A few properties of our data invite misreadings, or differ from what the
frozen texts literally say. They belong here, not in an appendix.

\begin{itemize}
\item \textbf{Step-0 twins are family-specific.} For some seeds the two
arms start from byte-identical step-0 rollouts, which reduces paired
variance. The twin sets differ by family: seeds 0--3 in family A (seed 4
forks), seeds 0 and 2--4 in family B (seed 1 forks), all five in family C.
Across seeds, rollouts are distinct in every family, so the five seeds are
genuine replicates; paired inference leans on twin structure only where it
exists.
\item \textbf{A conditioning trap in the rollout schema.} The leaky arm's
plus-pass rate \emph{conditional on base-pass} reads about 0.82 in every
family (\condplusA/\condplusB/\condplusC), which makes the leaky arm look
healthy on the extra tests. The comparable unconditional quantity ---
official base-and-extra (union) scoring computed over all rollouts ---
is slightly \emph{below} the hardened
arm's extra-pass rate in all three families (\uncondlpA\ vs \uncondhA;
\uncondlpB\ vs \uncondhB; \uncondlpC\ vs \uncondhC). Note the two columns
are different quantities: the leaky arm's is base$\wedge$extra while the
hardened arm's is extra-only, one more face of the disjointness defined
in \S\ref{sec:arms}. Our analyses use the
unconditional quantity; its direction independently corroborates
\S\ref{sec:avg}.
\item \textbf{Naive counting doubles the headline.} Counting every
generous failure as wrong code, with no artifact exclusion, roughly
doubles the verified-wrong counts (family A \naiveinclA\ vs our
\vwcountvfive; B \naiveinclB\ vs \famBvwcountvfive; C \naiveinclC\ vs
\famCvwcountvfive). Every verified-wrong number in this paper is the
artifact-excluded count.
\item \textbf{Small print.} The tracking analyses pair tasks with at least
10 rewarded rollouts (a threshold ported verbatim from the family-A
reference implementation; conclusions are unchanged under a
$>0$ threshold). The merged family-A classification file carries no seed
field, and a naive (task, step, uid) join leaves two of its \vwnvfive\
records seed-ambiguous; all per-seed numbers therefore use the
rollout-level re-join, which carries seeds exactly and reproduces the
pooled total with zero ambiguity (Appendix~\ref{app:vw}). The archived
rollout schema is
asymmetric across arms (documented in the data release), and the heavy FP
channels persist across seeds: \npersistent\ tasks appear in every seed's
top-twenty FP list, \npersistentrobust\ of them robustly to ties at the
twentieth rank.
\end{itemize}

\subsection{Statistical discipline}
\label{sec:discipline}

Three rules apply everywhere. First, the seed is the unit of inference:
with \nseeds\ seeds per arm, seed-level $t$ intervals carry four degrees
of freedom, and the exact paired sign-flip test cannot go below
$p=\pexactfloor$; we say so wherever such a $p$-value appears, so a floor
is never read as strong evidence. Second, no single-threshold verdicts:
preregistered decisions report their margin and the full interval,
exploratory results report intervals or sensitivity surfaces, and no
conclusion rests on one number crossing one line. Third, unless stated
otherwise, intervals are two-sided 90\% (equivalently, one-sided 95\%
against a margin), and bootstrap intervals state their resampling unit.

\section{Results}
\label{sec:results}

\subsection{Average held-out effect}
\label{sec:avg}

The preregistered primary question is confirmatory \regC: does training
against the leaky suite cost held-out ability? We test it as a one-sided
non-inferiority contrast \citep{schuirmann1987tost,lakens2017equivalence}.
With the gap defined as hardened-minus-leaky mean pass rate over the
\nevalpaired\ paired held-out tasks, the inferiority hypothesis (gap
$\geq \Xmargin$\,pt) is rejected when the one-sided 95\% upper confidence
bound on the gap falls below the margin; the margin was fixed before any
pilot data.

Both clauses of the preregistered benign verdict hold. At checkpoint
\nsteps\ the gap is \pgapmean\,pt (CI90 \pgapci), and its one-sided 95\%
upper bound is \pgapupper\,pt --- half the margin --- staying below
\Xmargin\,pt under all six estimator variants (uppers
\pgapsurfacelo--\pgapsurfacehi\,pt across percentile, basic, BCa,
hierarchical, seed-$t$, and sign-flip constructions). The second clause is
on the training side: the leaky arm's FP mass does not trend upward
(quarter means \fpquarterfirst\% $\to$ \fpquarterlast\%, trend CI90
\fptrendci\ includes zero). Meanwhile both arms improve over their base
model (+\learnmbpprange\,pt on held-out MBPP, +\learnherange\,pt on
HumanEval+), so the bounded gap is not two arms failing together.

We state what the design can and cannot see. With \nseeds\ seeds the CI90
half-width on the gap is \mdehalfboot--\mdehalft\,pt, so true gaps between
roughly 0.2 and \Xmargin\,pt are not reliably detectable. The result is a
bounded-effect statement --- this deployed leak sits below \Xmargin\,pt at
1--1.5B and \nsteps\ steps --- not equivalence at zero, and not a clearance
for longer runs or larger models (\S\ref{sec:limitations}).

\subsection{Leakiness tracking}
\label{sec:track}

\begin{figure}[t]
\centering
\includegraphics[width=\textwidth]{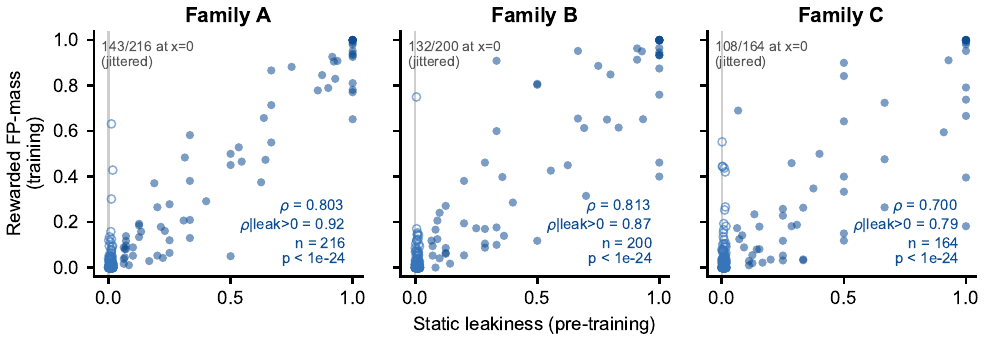}
\caption{\textbf{Rewarded false-positive mass tracks static leakiness in
all three families.} Per-task rewarded-FP mass during leaky-arm training
($y$) against the static leakiness score measured before training ($x$).
Each point is one task; zero-leakiness tasks are hollow and
right-jittered for visibility only (all statistics use the exact zeros).
(a)~Family A (primary): $n=\ntrackedtasks$, Spearman $\rho=\figrhoA$;
restricted to tasks with nonzero leakiness, $\rho=\figrhoAcond$
($n=\fignAcond$). (b)~Family B: $n=\fignB$, $\rho=\figrhoB$; nonzero
subset $\rho=\figrhoBcond$ ($n=\fignBcond$). (c)~Family C: $n=\fignC$,
$\rho=\figrhoC$; nonzero subset $\rho=\figrhoCcond$ ($n=\fignCcond$).
All six correlations have $p<\figpceil$. About two thirds of tasks sit
at zero measured leakiness (\figzeroA/\ntrackedtasks, \figzeroB/\fignB,
\figzeroC/\fignC); the tracking is not an artifact of that column ---
it strengthens on the continuous range. Rank ordering is consistent
across families; magnitudes differ, and family C also has the smallest
continuous-range sample.}
\label{fig:p2-scatter}
\end{figure}

Where the leaked reward goes is confirmatory \regC, and a cheap audit
predicts it. Per task, rewarded-FP mass during training tracks the static
leakiness score computed before training: Spearman $\rho=\rhotrackraw$ raw
(task-level bootstrap-95 \rhotrackbootci, \ntrackedtasks\ tasks), and
$\rho=\rhotrackpartial$ after controlling for task difficulty. An OLS
decomposition attributes the relationship to leakiness (coefficient
\olsbetaleak) rather than difficulty (\olsbetadiff), and excluding tasks
with few rewarded rollouts leaves the correlation essentially unchanged.
We report rank correlations throughout; the task distribution is
zero-inflated (most tasks have no measurable leak), which product-moment
correlations overstate. For scale: the pooled static FP rate of the leaky
suite is \staticrawmass\% raw, \staticexclmass\% artifact-excluded.
Figure~\ref{fig:p2-scatter} shows the per-task relationship in all three
families.\footnote{Three zero-leakiness tasks (one per family:
\outtaskA, \outtaskB, \outtaskC) show rewarded-FP mass above one half.
These are not easy tasks the model already solves; they are tasks the
base model rarely passes under the original suite (1, 4, and 1 of the
$k{=}16$ audit samples), so each ``zero'' rests on at most four
observations, and zero observed false positives among four passing
samples is consistent with true leakiness as high as \outubfour\ (exact
binomial 95\% upper bound; \outubone\ for the single-pass tasks). We
read these points as the finite-sample floor of the static audit on
hard tasks, not as evidence of training creating false positives on
genuinely non-leaky tasks. Per-task receipts are in the released notes,
including a buffer band below one half where one task (\outtaskband)
exceeds its static upper bound --- the audit measures the base policy,
and the tracking claim is rank tracking, not per-task calibration.}

The reading this supports is inheritance, not instruction. The weak suite
does not teach the model wrong programs; it fails to correct the wrong
programs the model already produces, exactly where the static audit says
it will fail to correct them. Section~\ref{sec:mech} tests that reading
directly.

\subsection{Verified-wrong: what the leaked reward paid for}
\label{sec:vw}

FP mass says reward leaked; it does not say what the reward paid for. This
section is exploratory \regE. A rewarded FP can contain a genuinely wrong
program, or correct-enough code that a defective plus test rejects, an
input that violates the task's contract, or a harness timeout. The
exhaustive audit of \S\ref{sec:vwspec} separates these for every one of
the \vwnvfive\ family-A records.

Close to half of the rewarded FP mass is verified genuinely wrong code.
The record-weighted point estimate is \vwsharevfive\%
(\vwcountvfive/\vwnvfive), with task-cluster bootstrap-95 interval
\vwclusterci: the interval is wide because FP records cluster by task, and
we quote it wherever the point estimate appears rather than the tighter
(and wrong) naive binomial. Estimator variants stay within a two-point
band of the point estimate (Table~\ref{tab:vw}), and \vwanyvwvfive\% of
audited tasks contain at least one verified-wrong record. Under the claim
ladder pre-written in the registration (\S\ref{sec:prereg}), this selects
the strongest of the three pre-committed sentences: the reward stream
exploits genuine false positives of the deployed suite --- where
``exploits'' means collects reward from, and \S\ref{sec:mech} shows the
collection is selection rather than learning. Every row of
Table~\ref{tab:vw} also sits far above the \stoprule\% floor at which the
registration would have downgraded the claim.

\begin{table}[t]
\centering
\caption{Verified-wrong share of rewarded FP mass \regE. Family A rows are
estimator variants of one quantity under the v5 specification; the
mandatory interval is the task-cluster bootstrap. B/C intervals are seed
$t$-CI90 (different construction; stated per row). Cluster-bootstrap
endpoints carry Monte-Carlo tolerance of roughly $\pm$1\,pt.}
\label{tab:vw}
\begin{tabular}{lll}
\toprule
Estimator & Share & Interval \\
\midrule
Family A, record-weighted (headline) & \vwsharevfive\% & \vwclusterci\ (cluster b.-95) \\
\quad code-deduplicated & \vwcodededupvfive\% & (\vwcodededupcount/\vwcodededupn) \\
\quad task-unweighted mean & \vwtaskmeanvfive & \vwtaskmeanci \\
\quad unresolved-excluded & \vwunresexclvfive\% & --- \\
\quad generous-timeout forced-fail & \vwforcedfalsevfive\% & --- \\
\quad early / late training half & \vwearlyvfive\% / \vwlatevfive\% & --- \\
\quad excluding Mbpp/781 & \vwsharevfiveexclA\% & (1340/2743) \\
Family B (deepseek-coder-1.3b) & \famBvwsharevfive\% & \famBseedci\ (seed $t$-90) \\
Family C (Llama-3.2-1B) & \famCvwsharevfive\% & \famCseedci\ (seed $t$-90) \\
\bottomrule
\end{tabular}
\end{table}

The audit is self-penalizing, and we report its trail. The share fell from
\vwshare\% under the first signed specification to \vwsharevthree\% and
then \vwsharevfive\% as adjudicated amnesties accumulated; every signed
predicate could only move the number down, which is why we trust the
endpoint more than the earlier, larger values. Both replication families
reproduce a large residual share (Table~\ref{tab:vw}); family C's larger
figure is the least-audited of the three and carries a composition caveat
(\S\ref{sec:xfam}), so the three families replicate ``large,'' not one
number.

Accounting, with units spelled out. The leaky arm banks \rewardgap\,pt
more train reward than the hardened arm, and FPs are \fpmassshare\% of its
rewarded rollouts (a percent share, not points). The non-verified-wrong
remainder of FP mass splits across the four non-model causes of
\S\ref{sec:vwspec}; Mbpp/267, 639, 771, and 806 are worked examples in the
appendix. Counting naively --- any generous failure, no artifact exclusion
--- roughly doubles every count in Table~\ref{tab:vw}
(\S\ref{sec:honesty}).

\subsection{Stratification: exposure on the train side, association on the eval side}
\label{sec:strat}

The train side is confirmatory \regC. The registered P3 test stratifies
the \ntrain\ training tasks by the static audit (\ptrainnleak\ leak-stratum
tasks, \ptrainnclean\ clean) and compares the leaky arm's FP share across
strata. The gap is large and stable: \ptrainleak\% of rewarded rollouts on
the leak stratum are FPs, against \ptrainclean\% on clean tasks (both
means over \nseeds\ seeds) --- a difference of \ptrainshare\,pt (seed
$t$-CI90 \ptrainci). Like every
FP-mass quantity, this is classifier-free (\S\ref{sec:evalproto}), so the
audit amnesties cannot move it.

The eval side is exploratory \regE\ --- permanently. The registered P3
concerned the train side; the stratification below was computed after
seeing the data, and no later evidence upgrades its register. On the 47
held-out tasks with nonzero static leakiness (``exposed''), the hardened
arm ends \ptrainstrat\,pt ahead of the leaky arm --- positive means the
leaky arm did worse --- with seed-level $t$-CI90 \pstratci\ and
\pstratseeds\ seeds positive, against \pcleanstrat\,pt on the 81 clean
tasks. The difference-in-differences is \pdid\,pt (one-sided seed-level
$p=\pdidp$; the exact test floors at \pexactfloor). Within the exposed
stratum: tasks the base model can already sometimes solve show
\pfloorthirty\,pt, the mid-leakiness band \pmidband\,pt, and the
high-leakiness band \phighleak\,pt --- the last is a measurement floor,
not an absence: \highleakbasezero\ of those 20 tasks have base pass rate
exactly zero, so there is no improvement to lose.

Four riders are mandatory wherever this result is used. First,
concentration: the top five carrier tasks hold \pconc\% of the
exposed-stratum summed gap, and the seed-level interval crosses zero when
they are left out; dropping single seeds (jackknife) also crosses zero.
Second, specification search: we priced our own analysis freedom by
embedding the exposed-stratum test in a family of \ssnvalid\ valid
specifications. The chosen specification ranks \ssrankT\ of \ssnvalid\ by
$t$-statistic and \ssrankGap\ by effect size; family-adjusted $p$-values
are \ssNaFamily\ and \ssNbFamily\ under two arm-relabel nulls and
\ssNtwoFamily\ under a covariate-shuffle null. The audit prices one axis
of freedom (the stratum boundary), not all of them (estimand, inference
unit, checkpoint), so the adjusted values are lower bounds on the true
correction; we report the surface and attach no summary adjective. Third,
no causal sentence: the design that could separate ``the leak damages
these tasks'' from ``these tasks attract FPs for other reasons'' is
blocked on this grid (\S\ref{sec:limitations}). Fourth, the apparent
deepening of the exposed gap across checkpoints is a descriptive
aggregate; per seed it is monotone in one of five.

\subsection{Mechanism: selection of pre-existing error modes}
\label{sec:mech}

This section is exploratory \regE. The reward stream demonstrably contains
a channel that favors wrong code: in \fponlyA\% / \fponlyB\% / \fponlyC\%
of GRPO groups containing at least one base-suite-passing completion
(families A/B/C; family-A seed $t$-CI95 \fponlyAci), every such completion
is a false positive, so the group's entire positive advantage flows to
wrong programs. The mechanism question is
whether training climbs this channel --- learned exploitation --- or
merely samples it --- selection.

Climbing would show up as growth, and we do not detect any. Rewarded-FP
mass is flat across training in families A and C and declines in family B
(trend CI90 \fpdeclineBci); the FP-only group share does not grow
(last-quarter minus first-quarter intervals include zero in all three
families); and the per-task FP pattern is in place early and stays put
(Spearman with static leakiness \rhoearly\ in the early window, \rholate\
late, per-task growth $\approx 0$). These are not-detected results at
$n=\nseeds$, not proofs of zero (\S\ref{sec:discipline}); a preregistered
800-step continuation sharpens them (\S\ref{sec:cont800}).\footnote{Two
post-hoc robustness checks (not preregistered) back this up.
Re-estimating the trend on the genuine-wrong stratum alone (task fixed
effects within seed, seed-clustered CI90) leaves every per-family slope
in [\genslopelo, \genslopehi]\,pp per 100 steps, with a worst-case CI90
upper bound of \genworstub --- at most about +\genmaxrise\,pp over the
full run against a $\sim$\genlevel\,pp level; excluding the \npredtasks\
signed defect-predicate tasks, or thresholding by static leakiness,
changes nothing. The flatness is therefore not an artifact of pooling
defect mass with genuine-wrong mass. Separately, the mild downward drift
of the genuine share in two of three families is consistent with passive
dilution by the rising base-pass rate (positive in \basepassriseall\
seeds); we find no evidence that training teaches avoidance of genuine
FPs, and power is insufficient to exclude a cleaning component --- we
commit to neither reading.} Nor does the
paid-for code consolidate: on the largest persistent channel, all
\distinctlargest\ verified-wrong records are distinct programs --- the
pool as a whole does contain duplicates, so zero repeats is informative
--- and a targeted scan across all three families' FP pools finds no
hacking signatures. Distinctness refutes fixation, not learning in
general.

The decisive test samples the initialization. We generated fresh
completions from the three untrained base models on the persistent
channels, applied the same leaky filter (passes MBPP, fails MBPP+), and
asked whether each channel's characteristic wrong output is already
there. It is: the base FP subset reproduces the channel's wrong output at
or above the trained rate on \basectrlfam\ of \basectrlchannels\ channels
(family-mean; \basectrlpooled\ pooled across families, \basectrlguard\
under a minimum-denominator guard). Unconditioned base samples sit below
trained rates on every channel --- the leaky filter, not training, does
the selecting. What training changes is the frequency of these outputs
within the accepted set, not the shape of what is wrong.

The advantage structure explains why selection is available where
steering is not. Where FPs and genuine passes coexist in a group, both
earn identical reward, so the group-relative advantage carries no signal
that separates them; the only all-FP advantage flows are the FP-only
groups above. On fully leaky tasks, \lfourfp\% of rewarded completions in
mixed-reward groups are FPs, and only \lfourhonestleak\% of those groups
contain even one honest solution to contrast against, versus
\lfourhonestclean\% on clean tasks.\footnote{Denominators, pooled over
\nseeds\ seeds: \lfourdenomcompl\ base-pass completions in fully-leaky
mixed groups; \lfourdenomleak\ fully-leaky and \lfourdenomclean\ clean
mixed groups. A \emph{mixed} group has 1--7 base-passers of 8. ``Clean''
here means static leakiness exactly zero; the 21 train tasks the static
probe could not score are excluded (folding them in shifts
\lfourhonestclean\% to \lfourcleannull\%). This is a stricter set than
the \ptrainnclean-task clean stratum of \S\ref{sec:strat}, which also
contains the unscored tasks.} One tempting simplification fails,
and we flag it as falsified rather than dropping it quietly: fully leaky
tasks are \emph{not} saturated zero-gradient pockets --- \zerogradmixed\%
of their reward groups have mixed rewards (independently cross-checked).
All of this cashes out as measurement inflation without a capability
trace: the leaky arm banks \rewardgap\,pt more train reward, while the
held-out gap stays at \pgapmean\,pt (CI90 \pgapci; one-sided 95\% upper
bound \pgapupper\,pt, \S\ref{sec:avg}).

A weight-level worry survives the behavioral evidence: the model could be
sharpening toward its FPs below behavioral resolution. We tested the
readable half of this on families B and C (the two with retained final
checkpoints, \nseeds\ seeds each): per-token log-probability sharpening
of FP versus genuine completions, length-matched, because raw comparisons
are confounded (FP completions run \sharplenconf\ tokens longer). After
length correction there is no detectable preferential sharpening above
the design's \sharpfloorlo--\sharpfloorhi\ nats-per-token resolution
floor: family B is a clean null under all matchings, and family C is null
under overlap-only matching. This is a bounded null, not a proof;
parameter-level effects below the floor, longer horizons, and larger
models remain open (\S\ref{sec:limitations}).

The picture that survives every test we ran: a real, persistent
exploitation channel, visible to a cheap static audit before training,
whose contents predate training --- selected, not taught, within the
horizon we measure.

\subsection{Cross-family replication}
\label{sec:xfam}

The replication is confirmatory \regC. The registration froze three
criteria and keyed them to family B before any replication data existed: a
verified-wrong share of at least \stoprule\%, tracking of the static
audit, and the train-side stratification gap. All three pass on family B:
verified-wrong \famBvwsharevfive\% (seed $t$-CI90 \famBseedci), tracking
$\rho=\gatebC$ (CI90 \gatebCci), train-side gap \gatecC\,pt (CI90
\gatecCci). Family C shows the same three signatures: \famCvwsharevfive\%
(\famCseedci), $\rho=\famCctwo$ (\famCctwoci), and \famCcthree\,pt
(\famCcthreeci). Table~\ref{tab:xfam} lines the families up.

\begin{table}[t]
\centering
\caption{Coarse-axis concordance across model families. Interval
constructions differ by cell and are stated inline; family A's
verified-wrong interval is the task-cluster bootstrap, B/C are seed
$t$-CI90. The transfer claim is locked to the first three rows; held-out
task-level effects do not transfer (see text). Family C's share is the
least audited of the three (\S\ref{sec:limitations}).}
\label{tab:xfam}
\footnotesize
\setlength{\tabcolsep}{2pt}
\begin{tabular}{lccc}
\toprule
 & A (Qwen-1.5B) & B (dsc-1.3b) & C (Llama-1B) \\
\midrule
Tracking $\rho$ \regC & \figrhoA\ \rhotrackbootci & \figrhoB\ \gatebCci & \figrhoC\ \famCctwoci \\
Train-side gap, pt \regC & \ptrainshare\ \ptrainci & \gatecC\ \gatecCci & \famCcthree\ \famCcthreeci \\
Verified-wrong share \regE & \vwsharevfive\% \vwclusterci & \famBvwsharevfive\% \famBseedci & \famCvwsharevfive\% \famCseedci \\
Eval-side exposed gap \regE & \ptrainstrat\,pt (assoc.) & \cfourB\,pt (reversed cand.) & positive, underpowered \\
\bottomrule
\end{tabular}
\end{table}

What transfers is exactly this coarse axis --- the failure-cause taxonomy
and the two train-side regularities --- and nothing finer. The adverse
facts belong next to the concordance, so here they are. Held-out
task-level effects do not transfer: the correlation between families B
and C's per-task oracle deltas is \bcTaskDelta\ (negative; the arm delta
itself is near zero, so a near-zero-region correlation carries little
information, but we state it rather than hide it), and \zeroleakybetter\
of \nevalmbpp\ eval tasks favor the leaky arm in all three families.
Family B's eval-side exposed-stratum gap runs opposite to family A's:
\cfourB\,pt (seed-CI90 \cfourBci), a reversed-direction candidate signal,
not confirmed by the exact small-$n$ test ($p\approx\cfourBp$, floor
\pexactfloor); family C is directionally positive but underpowered
($p=\cfourCp$). The eval-side association of \S\ref{sec:strat} therefore
does not replicate cleanly, which is one reason it stays exploratory.
Family B's average-effect verdict is mixed rather than benign: the only
failed benign clause is its FP-mass trend, whose CI90 \fpdeclineBci\
excludes zero in the negative --- anti-exploitation --- direction. And
family C's \famCvwsharevfive\% is the least-audited share of the three,
with \famCsyntaxfail\% syntax-invalid extracted code among its failures
(untreated composition; \S\ref{sec:limitations}) --- it should not be
read as ``family C is harmed most.''

Finally, a preregistered symmetry check on family B's evaluation failures
finds no exposure-specific asymmetry under its primary reading
(profile-divergence delta \rthreedeltaexposed, $p=\rthreep$); a
sensitivity reading is positive in all five seeds (\rthreesens, bootstrap
CI90 \rthreesensci, not excluding zero) and we report it as borderline.
Per the wording map pre-stated in the protocol, the association reading
of \S\ref{sec:strat} stands unchanged.

\subsection{An 800-step continuation}
\label{sec:cont800}

The strongest objection to \S\ref{sec:mech} is horizon: 400 steps might
be too early. We tested this with a confirmatory \regC\
continuation, preregistered before its data existed: fresh
\contsteps-step runs of both arms in families B and C, \contseeds\ seeds
each, with the learning-rate schedule verified to span the full
\contsteps\ steps in all \contruns\ runs. The frozen criterion for
growth had three parts --- a positive FP-only-share slope excluding
zero, a second half above the first, and at least one secondary measure
agreeing --- and none fired. The FP-only share slope excludes zero on
the positive side in none of the \contruns\ runs (family aggregates
\contBslope\ \contBslopeci\ and \contCslope\ \contCslopeci\ pp per 100
steps); the second half is not above the first (\contBhh\ \contBhhci;
\contChh\ \contChhci); and the secondary rewarded-FP mass trend also
spans zero (\contBmass, \contCmass). Under the preregistration's
pre-authorized wording, growth is refuted at twice the horizon, and
persistence survives.

The evidence is not symmetric between the families, and we state it the
way the power analysis does. Family B is adequately powered: its
minimum detectable slope at 80\% power is \contMDEB\,pp per 100 steps
against an observed slope near zero, so its null is a genuine
refutation at that resolution. Family C is an underpowered
non-detection: its MDE is \contMDEC\,pp per 100 steps, and a rise of
about four points over the full run would be missed with roughly
\contCmiss\% probability. Nor does the flatness mean a dead run: genuine-pass
rates rise throughout in both families (\contBtp\ \contBtpci;
\contCtp\ \contCtpci), and held-out accuracy rises with them. Two
disclosures. First, family C's unconditional rewarded-FP \emph{rate}
(not share) rises over the full window under one estimator (\contCrate\
\contCratecifew; the OLS interval \contCrateciols\ spans zero), driven
by the early window and confounded with the rising base-pass rate ---
consistent with early lock-in, not late growth. Second, the
continuation reuses the same \ntrain\ training tasks, so it refreshes
the seed axis only; the task axis is untested.
Table~\ref{tab:app-cont800} gives the per-run receipts.

A reclassification of the continuation rollouts under the frozen v5
predicates closes the remaining aperture. The apparent decline in family
C's genuine-FP share is directionally real \regC: all three continuation
seeds are negative, and the aggregate slope of \vfiveCagg\,pp per 100
steps (seed-$t$ CI90 \vfiveCaggci) excludes zero, firing the
preregistration's escalation clause for direction --- though with three
seeds the sign test bottoms out at $p=1/8$, weaker than the original
five-seed reading. But the decline concentrates in the early window
(\vfiveCearly\ \vfiveCearlyci), exactly where the base pass rate is
rising (\vfivebaseearly\,pp per 100 steps), and the dilution term alone
(\vfivedilution) covers the observed early drop. The late-window
genuine-FP \emph{emission rate} spans zero (\vfiveCghci) with a point
estimate shrunk from the \vfiveoldg\ anchor to \vfiveCgh, so the
preregistered demotion clause closes the active-cleaning reading: the
share falls because capability rises and dilutes the denominator, not
because training removes false positives.\footnote{Task-novelty
sensitivity: the continuation grids contain \vfivenovelB\ (family B) and
\vfivenovelC\ (family C) tasks absent from the original classification
set, holding \vfivenovelBshare\% and \vfivenovelCshare\% of records;
excluding them leaves every verdict in this paragraph unchanged.} No
continuation seed reproduces the earlier heavy tail (no seed with
$|\text{slope}|>1$); the tail-excluded anchor of \vfiveanchor\ replicates
at \vfiverepl. Family B's share stays flat (\vfiveBflat\ \vfiveBflatci),
reproducing its null.


\section{Can the judges audit themselves? A meta-audit}
\label{sec:meta}

\textbf{This section is exploratory \regE, and we fence it off before we
make it.} None of it entered the preregistration that governs
\S\ref{sec:results}; no result above depends on it; and it is not part of
the paper's contribution. It is a follow-up audit of the frontier
\emph{judges} themselves --- two subjects on MBPP under one frozen
instrument --- so it licenses nothing about ``frontier models'' in
general, and every number below is descriptive. Its discipline matches the
rest of the paper's: each of the program's five studies was preregistered
and frozen before any data existed, each carries a signed adversarial chair
ruling on top of a standing three-lane audit, no kill-finding issued in any
of the five, and in two of them the chair reproduced the headline
statistics bit-exact by an independent route --- for the attribution test, a
baseline-free route that proves the reported contrasts are invariant to the
silent calibration baseline.

Two of the program's contrasts were confirmatory under their own separate,
later preregistrations (flagged where they occur); that is a property of
those preregistrations, not of this paper. Unless stated otherwise,
intervals in this section are task-cluster bootstrap 95\% CIs
($B{=}10{,}000$, resampling unit $=$ task cluster), stated as 95\%
throughout to separate them from the paper's two-sided-90\% default
(\S\ref{sec:discipline}). One naming caution: the judge subjects here are
the frontier-lab models \emph{DeepSeek-V4-Pro} and \emph{MiMo-v2.5-Pro}
(with \emph{GPT-5.6} as a witness reader)\footnote{The witness ran in the
vendor's desktop client (session telemetry originator ``Codex Desktop'',
v0.144.0-alpha.4; the client ships within the ChatGPT desktop application)
as model \texttt{gpt-5.6-sol} at reasoning effort \texttt{ultra};
session-level telemetry (\texttt{thread\_settings\_applied}) confirms this
configuration for all 35 judging batches with no deviation (receipt summary:
\texttt{runs/metaA\_dual\_role/gpt\_lane/RECEIPTS\_gpt\_witness\_config.md}
in the repository). The \texttt{ultra} configuration permits automatic
sub-task delegation within the client, so ``GPT-5.6'' in this paper denotes
this configured system as deployed, not a bare single-model forward pass.};
\emph{DeepSeek-V4-Pro} is not the
trained family-B subject \famBmodel\ of \S\ref{sec:xfam}, despite the shared
vendor.

\regE\ As a preregistered insurance probe, two frontier judges reading
the code blind and without execution separated the rewarded
verified-wrong false positives from genuine passes at AUC
\fjgptB/\fjgptC\ (GPT-5.6, families B/C) and \fjdsB/\fjdsC\
(DeepSeek-V4-Pro), decisively above a code-length baseline
(\fjlenB/\fjlenC; task-clustered bootstrap 95\% CIs non-overlapping),
establishing that separability exists above our 1.5B policy's
capability --- a read-only discrimination result, not a claim about
frontier behavior under reward pressure.\footnote{Contamination
disclosure (preregistered at freeze, before any judging ran): GPT-5.6
had performed an unrelated review of this repository on 7/10; the six
tasks it demonstrably inspected were excluded from the blind sample,
but residual model-side familiarity with the codebase cannot be ruled
out or measured. The DeepSeek-V4-Pro lane had no repository exposure;
its AUC (\fjdsB/\fjdsC) is therefore the contamination-free floor, and
the conclusion does not rest on the GPT-5.6 lane.} That probe raises the
follow-through question this section answers only weakly: can those same
frontier judges separate their \emph{own} false positives? We ran it as a
dual-role audit --- each judge scoring code drawn from its own output
distribution --- on two subjects, deconfounded it by cross-judging, and
stress-tested it with explicit authorship claims.

\textbf{The gap, as a difference between settings.}\quad Before any
number: the self-assessment score and the hetero-judge anchor are measured
on \emph{different item pools} --- a judge's own false positives versus the
1.5B policy's false positives --- so the gap between them is a difference
between two settings, not an estimate of a self-vs-hetero effect, and no
causal reading attaches to it. Within that limit, both subjects separate
their own false positives from genuine passes only weakly, descriptively
below their own hetero-judge anchor. On its own-distribution errors,
DeepSeek-V4-Pro's self-assessment discriminates FPs from genuine passes
weakly but above chance (AUC \maSelfDS\ \maSelfDSci), descriptively far
below the \maAnchDS\ hetero-judge anchor measured on a different
(1.5B-policy) FP pool; the item pools differ, so the gap is not an
estimate of a self-vs-hetero effect. MiMo-v2.5-Pro self-assesses its own
FPs weakly but above chance (AUC \maSelfMM\ \maSelfMMci), descriptively
below its own \maAnchMM\ \maAnchMMci\ hetero-judge anchor (different item
pool); two-sample difference CI \maDiffMMcilic, excludes 0; not a
self-vs-hetero effect estimate. Two further limits travel with these
numbers. Most of the residual signal is between-task
(\maBtwDS\%/\maBtwMM\% of FP rows have no within-task genuine comparator),
so the AUCs may substantially reflect task-level leakiness recognition
rather than per-sample detection of a judge's own errors; and per-task
own-FP rate tracks the paper's static leakiness table (exploratory
$\rho{=}\maRhoDS/\maRhoMM$). The MiMo arm is the weaker of the two: its
earlier ``non-overlap margin'' is retired as convention-sensitive, so its
load rests on the difference CI above --- not the same evidential comfort as
the DeepSeek arm, whose own gap additionally keeps a conservative
non-overlap slack (self CI-high below anchor CI-low, margin .168; DeepSeek
only).\footnote{The two subjects were generated under an intent-equalized
budget, not a nominally identical one: MiMo ran at \texttt{max\_tokens} 4096
vs DeepSeek's 2048, because MiMo's hidden-reasoning tokens count against its
completion cap (at 2048, 32.5\%, i.e.\ 13/40, of MiMo pilot generations were
empty vs DeepSeek's 2.8\%; the 4096 pilot gate read 0.0\% empty, $n{=}40$).
Base-pass rates therefore compare equalized-by-intent instruments. The full
cross-subject receipts, the diagnostic $p_{\mathrm{pass}}{<}0.5$ flag counts
(which are non-preregistered, descriptive, and are judge house-style rather
than any self-leniency), and the licensed long-form readings are in
App.~\ref{app:meta}.}

\textbf{Deconfounding by cross-judging: unresolved.}\quad Crossing the two
judges over the two locked pools removes the item-pool confound within each
contrast, and the large gaps do not reappear as self-vs-hetero differences
on identical items. The preregistered confirmatory contrast (its own,
separate prereg) is $\Delta_{\mathrm{DS}}{=}\maDeltaDS$ \maDeltaDSci, which
covers zero; its frozen ruling is \textsc{unresolved}. There is no evidence
at this resolution that either author ranks its own rows worse than the
cross reader; equally, a same-author deficit is not excluded, since a
penalty cancelled by an opposing reader-capability difference remains an
un-excluded explanation. Each $\Delta$ covers 0; the point estimates sit
slightly below the author ($\Delta_{\mathrm{MiMo}}{=}\maDeltaMM$
\maDeltaMMci). All four judge$\times$pool cells sit in a narrow band
(\maBand), descriptively far below the same judges' 1.5B-pool anchors
(\maAnchMM/\maAnchDS): the band follows the pool, not the reader and not
authorship. Both within-task contrasts cover zero, so no per-sample
detection reading is licensed for any cell. A secondary summed
$2{\times}2$ interaction contrast is reported symmetrically and licenses no
additivity-model conclusion in either direction; it, the within-task
readouts, and a $400$-row falsification reference are reported only in
App.~\ref{app:meta}. Register note: to our knowledge, this is the
pre-registered first deconfounded test of the same-author question; its
confirmatory contrast returned UNRESOLVED (neither a same-author deficit
nor its absence is established). Both directions were declared publishable
at freeze.

\textbf{Explicit authorship claims.}\quad A separate design tells the judge,
in text, that the code is its own or another model's, and measures the
interaction. A stateless API cannot verify such a claim, so the manipulated
variable is claim \emph{text}, not knowledge; we use no
recognition/memory/knowing language. For the DeepSeek judge the confirmatory
interaction is $I(\mathrm{DS}){=}\maIDS$ \maIDSci, two-sided bootstrap
$p{=}\maIDSp$, Holm-surviving at the .025 threshold (family
$\{I(\mathrm{DS}),I(\mathrm{MiMo})\}$), in the harsher-when-truly-own
direction: within its own pool the claim pair moves the judge toward relative
harshness on its own false positives ($K{=}{-}.043$ $[-.089,\,+.001]$), and
on the other author's pool the same claim pair moves it toward relative
leniency ($K{=}{+}.063$ $[+.013,\,+.118]$, exploratory, outside the family);
the sign flip between the true-claim and false-claim configurations IS the
confirmatory interaction, and the full $2{\times}2$ is reported with no
single-story compression (App.~\ref{app:meta}). This pattern is consistent
with truth-sensitivity of the response through any channel (stylistic
familiarity explicitly among them) AND, inseparably, with a
claim-content~$\times$~pool-property interaction that is not truth-mediated
(pool difficulty or FP-type composition modulating susceptibility to
authorship framing); both readings are named wherever this result is quoted.
Not licensed: recognition/memory/knowing; channel attribution;
``deception''; unqualified ``the shift is truth-sensitive''. For the MiMo
judge there is no detected attribution effect of either kind at this
resolution ($I(\mathrm{MiMo}){=}\maIMM$ \maIMMci, $p{=}\maIMMp$, not
rejected at Holm .05; both $K$ contrasts cover 0) --- a non-detection, not
truth-invariance and not equivalence, and the both-judges compounding row
does not fire (the DS judge detected). The $(\mathrm{MiMo},\mathrm{MiMo})$
calibration cell drifted by itself, so all its per-cell $D$ readouts are
descriptive only; the interaction $I$ and the contrasts $K$ are unaffected,
being attributed-cells-only statistics in which the silent baseline cancels
row-wise. The chair recomputed every $K$ and both $I$ with that silent
calibration baseline deleted and reproduced all six to ten decimal places,
so the drift that downgrades the $(\mathrm{MiMo},\mathrm{MiMo})$ per-cell
readouts does not reach the confirmatory $I(\mathrm{DS})$; what the drift
bounds is resolution (drift\_rms $.156$--$.322$), not the null (full frozen
statement in App.~\ref{app:meta}). Detection status is reported per judge;
we do not compare $I$, $K$, or $D$ across the two judges, because attribution
sentences may parse differently per vendor.

\textbf{A witness reader: motivation, not result.}\quad A third reader was
probed as a witness only. By the design of its single blind session it
carries zero confirmatory weight, and no wording can lift that ceiling; we
state it before quoting anything, and keep every witness number in
App.~\ref{app:meta}. Descriptively, the ordering there is: the 1.5B policy's
false positives are read well by all three readers tested; the two frontier
subjects reading their own-distribution frontier false positives sit in the
weak band above (\maBand, above chance); and the highest-scoring reader we
tested (GPT-5.6, its known 7/10 repository exposure disclosed and residual
familiarity not ruled out) improves on that band in the one blind session
yet stays far below every 1.5B anchor. Both known bias sources --- that same
exposure plus within-run accumulation --- push the witness numbers upward,
in the judge's favor, so correcting the known biases could only lower them;
single-instance variance is unquantifiable in both directions, and no bound
follows. Read as motivation and not as a result, this is what a
capability-relative supervision picture would look like on this instrument:
cheap error legibility is not shown to scale for free with generator
strength. The two subjects are frontier-lab models, though below the
strongest public tier at the time of writing, so the weak band already
occurs at a generator step below the top tier and the concern it raises does
not rest on generators having reached any ceiling; these data are silent, in
both directions, about generators not tested (App.~\ref{app:meta}).

\textbf{What the meta-audit adds, and does not.}\quad Staying inside those
fences: as an empirical answer to whether a frontier judge hides its own
errors when grading its own output, no self-protection was detected in this
pipeline, and the one confirmatory effect --- the DeepSeek attribution
interaction --- points the other way, toward relative harshness when the work
is truthfully labeled its own, inside a disjunction that cannot attribute it
to any self mechanism. The deconfounded ranking test is \textsc{unresolved}.
The attribution interaction and the cross-judging contrast are different
statistics under different manipulations; sign agreement across them carries
no evidential weight and is not narrated as convergence, and the attribution
result does not address the unresolved ranking contrast. What the meta-audit
adds to the paper is a capability-relative reading of the insurance result:
the false positives a cheap suite pays for are readable well above the 1.5B
policy, yet sit in a weak band both for the frontier subjects that produce
them and for the highest-scoring reader we tested --- an observation we carry into
\S\ref{sec:discussion} as motivation, not a law, and bound in
\S\ref{sec:limitations}, and, as throughout, a statement about two subjects
on this instrument, not about frontier models. Table~\ref{tab:meta} collects
the five headline readouts; the per-study receipts, calibration arm, witness
protocol, and the full frozen readings are in App.~\ref{app:meta}.

\begin{table}[t]
\centering
\footnotesize
\setlength{\tabcolsep}{5pt}
\caption{Meta-audit headline readouts (exploratory \regE; descriptive).
Intervals are task-cluster bootstrap 95\% CIs. The self-assessment AUC and
the hetero anchor sit on \emph{different item pools} (a judge's own FPs vs
the 1.5B policy's FPs), so ``Self $-$ anchor'' is a difference between
settings, not a self-vs-hetero effect; no causal reading attaches. The
cross-judge $\Delta$ is on \emph{identical} items. ``Holm'' marks the one
confirmatory-under-its-own-prereg survivor; ``UNRESOLVED'' is the frozen
ruling on the deconfounded contrast. The MiMo Self $-$ anchor upper bound
prints as $-.058$ here (table convention) and $-.057$ in the licensed
sentence of App.~\ref{app:meta}; both are frozen. Subjects are frontier-lab
models below the strongest public tier at the time of writing. Full
receipts: App.~\ref{app:meta}.}
\label{tab:meta}
\begin{tabular}{lcc}
\toprule
Readout & DeepSeek-V4-Pro & MiMo-v2.5-Pro \\
\midrule
Self-assessment AUC (own FPs) \regE            & \maSelfDS\ \maSelfDSci   & \maSelfMM\ \maSelfMMci   \\
Hetero-judge anchor (1.5B-FP pool)             & \maAnchDS\ \maAnchDSci   & \maAnchMM\ \maAnchMMci   \\
Self $-$ anchor (two-sample cluster boot)      & \maDiffDS\ \maDiffDSci   & \maDiffMM\ \maDiffMMcitab \\
Cross-judge $\Delta$ (deconfounded, same items)& \maDeltaDS\ \maDeltaDSci\ (\textsc{unres.}) & \maDeltaMM\ \maDeltaMMci \\
Attribution interaction $I$ (claim text)       & \maIDS\ \maIDSci\ (Holm) & \maIMM\ \maIMMci \\
\bottomrule
\end{tabular}
\end{table}

\section{Discussion}
\label{sec:discussion}

\textbf{Where deployed suites sit on the noise map.} Theory gives our
training-side results a map. \citet{rad2026rateorfate} model RLVR under a
noisy verifier and find a phase transition in the verifier's Youden index
$J=\mathrm{TPR}-\mathrm{FPR}$: incorrect modes amplify when $J<0$ and die
out when $J>0$. A deployed-but-weak suite --- too accepting, but still
separating right from wrong on average --- sits well inside the $J>0$
region, and the prediction that matters there is negative: no
amplification. That is what we measure, in all three families
(\S\ref{sec:mech}). Family B goes further and drifts in the predicted
decay direction (FP-mass trend CI90 \fpdeclineBci). Families A and C do
not: their FP share is flat at our horizon and power --- persistence
rather than the theory's asymptotic die-out --- which we flag as an open
resolution question; a preregistered continuation at twice the horizon
extends the flat reading (\S\ref{sec:cont800}). What the aggregate register
cannot see at all is the structure below it: the same run that keeps the
held-out gap inside a preregistered margin (\S\ref{sec:avg}) pays close
to half of its leaked reward for real bugs (\S\ref{sec:vw}), concentrated
on statically predictable tasks. Rate-benign is not uniform, and averages
are the wrong resolution for reward QA.

\textbf{Audit economics.} The FP mass is very unevenly distributed: its
concentration Gini is \fpgini\ (computed over per-task rewarded-FP counts
on all \ntrain\ training tasks, zeros included), and repairing the top
\topfixn\ task suites removes about half of it --- with the caveat that
the top ranks include channels whose mass indicts the plus suite's own
tests rather than the model (\S\ref{sec:vw}), so ``repair'' means fixing
tests on both sides. The static audit can also be upgraded free of
charge: training rollouts act as natural mutants that strictly refine it,
flagging \mutextratasks\ tasks the static probe had called clean, with
zero reversals (exact one-sided McNemar, $p\approx\mcnemarp$); the static
zeros were undersampling (median \staticzeromedian\ base-passing
samples), not cleanliness. The refinement buys localization, not volume
--- total escape mass is unchanged (\escstat\% vs \escnat\%) --- which is
what a QA tool needs.

\textbf{What hardening buys.} Hardening the suite removes the measurement
inflation: the \rewardgap\,pt reward gap is the price of the leak, and
the hardened arm does not pay it. At this scale it buys little else ---
held-out ability moves by \pgapmean\,pt (CI90 \pgapci; \S\ref{sec:avg}).
The practical rule this leaves is about where to measure, not how hard to
score: the inflation lives in the tasks the model trained on and, held
out, stays inside the bound of \S\ref{sec:avg}, so reward QA has to be
computed on tasks the model has not trained on. Train-task reward is an
accounting of what was paid, not of what was learned.

\textbf{Why the same holes pay across architectures.} The base-model
control of \S\ref{sec:mech} found that three unrelated architectures
already produce the same characteristic wrong outputs before any
training, and that the leaky suite's role is to select them into reward.
That three base models share error modes is less surprising than it
looks: the Platonic Representation Hypothesis \citep{huh2024platonic}
argues that models trained on similar data converge toward a shared
statistical model of the world, and shared wrong outputs on shared tasks
are an output-level face of that convergence --- our static audit's
cross-family stability (per-task leakiness correlates
\staticxfamlo--\staticxfamhi\ between families) is consistent with the
same reading. We flag the scope honestly: this is a minority pattern
among shared-failure tasks, not the dominant mode, and we offer it as
context for the selection result, not as an independent claim. The
practical consequence is unpleasant: a suite hole aligned with a shared
prior is not one model's quirk. All three families here were paid for the
same wrong outputs on the overlapping channels, and fixing the suite once
removed that tax for every one of them. Audit results, likewise, transfer
across models better than damage estimates do (\S\ref{sec:xfam}).

\textbf{What clean instruments can and cannot certify.} Every aggregate
instrument we pointed at this system read clean: held-out averages inside
a preregistered margin, no measured growth, no sub-behavioral sharpening
above the resolution floor. Yet the channel is demonstrably present ---
paid for at \fpmassshare\% of leaky rollouts, close to half of it
genuinely wrong code. When nothing has climbed the channel, a null is the
correct reading; these instruments did not fail. The asymmetry is the
point: clean readings bound what happened, they do not certify what the
reward stream contains. Nor do interior instruments yet cover enough
of the computation to close the gap: even a frontier interpretability
lens reports covering only a small fraction of activation variance (the
Jacobian-lens workspace of \citet{gurnee2026workspace} accounts for no
more than 10\% at any layer), and its authors describe the tool as capturing
the underlying structure ``only approximately and incompletely.'' The
discriminant that worked here was cheap and external: audit the reward
interface itself --- static leakiness before training, a base-policy
control after --- rather than infer absence of structure from clean
dashboards. That is the sense in which this paper is a measurement
contribution: it makes a Goodhart gap visible at the reward interface ---
the one layer where, on every instrument we ran, it shows at all.

\textbf{Practitioner checklist.} Four actions follow directly from the
measurements. (1) Run the static audit before RLVR: one sampling pass
from the base model predicts where rewarded FP mass will land
($\rho=\figrhoA/\figrhoB/\figrhoC$ across three families). (2) Read
suite failures through the taxonomy before believing them: close to half
of FP mass was real bugs here, but naive counting doubles the number
(\S\ref{sec:honesty}), and part of what looks like model failure is the
extra tests rejecting correct code. (3) Do not read train-task reward as
capability: the \rewardgap\,pt inflation does not appear held-out within
the margin of \S\ref{sec:avg}; measure on tasks the model has not trained
on. (4) Harden for measurement, not for capability: repair is cheap where
mass concentrates (top \topfixn\ suites $\approx$ half), and the payoff
is trustworthy reward accounting, not a stronger model.

\textbf{Selection, not a learned confidence gap \regE}\quad A binary
suite reward assigns a false positive and a genuine pass the same
value. Under any RL that takes this suite verdict as its sole
correctness oracle the two are indistinguishable in the reward --- a
consequence of the verdict carrying no false-positive bit --- and under
GRPO specifically they receive identical group-normalized advantage.
The objective therefore provides no \emph{explicit} supervision
separating a model's own false positives from its genuine passes ---
within a group the two receive identical scalar advantage --- though
token-level gradients still differ across sequences; this specializes
the known
under-specification of binary rewards
\citep{kalai2025hallucinate,damani2025rewards} to the leaky-suite,
natural-false-positive case. The measurements are consistent with this:
the trained$-$base difference is negligible ($\leq\probemaxdiff$ nats
per token, on the log-probability readout), and whatever weak separation
exists is already present in the untrained
base, too small to act on as a false-positive interceptor (matched AUC
\probeauclo--\probeauchi\ on the primary readout; post-hoc, not
preregistered; within the \probelmretB\%/\probelmretC\% of paired-cell
records retained by length matching, with no extrapolation to discarded
cells). Notably, although RL reinforces the false positives
(they earn $+1$), it does not build a usable confidence gap over base:
on the preregistered primary readout the trained$-$base change includes
zero in both families under both the raw and length-matched estimators
(raw: \probetbBprim; \probetbCprim), and the four
sensitivity readouts whose CI90 excludes zero (\probetbBrb\ and
\probetbBrtwo\ in family B, \probetbCrb\ and \probetbCrtwo\ in family C,
nats per token) are small, mixed in direction, and not stable across
readouts --- the largest, \probetbCrtwo, flips sign to \probetbCrtwolm\
under length matching, and \probetbBrb\ is an erosion-direction reading
--- too small in every case to act on. Read this way, the false positives the leaky suite pays for
are honest errors rather than deception, reinforcing the
selection-not-learning account.
\regE\ Whether the frontier judges that catch the 1.5B policy's leaked
false positives can catch their \emph{own} is a separate, exploratory
question, and we keep it out of this paper's claims; the preregistered
insurance probe that motivates it, and the meta-audit that follows it, are
in \S\ref{sec:meta}.

\textbf{A recursive implication for audits.}\quad Our witness observation
carries a recursive implication for audit methodology itself, including
the methodology of this paper. If error legibility falls as generator
strength rises (\S\ref{sec:meta} --- motivation, not a result), then the
least legible errors in any pipeline are those produced by its strongest
models --- which are, increasingly, the same models employed as its
reviewers and auditors, with no stronger reader available above them. In
the one blind session we ran, even the highest-scoring reader we tested
stayed far below its 1.5B-pool anchor on frontier-generated pools; to the
extent that pattern holds, oversight strategies premised on finding a
stronger judge would not extend past the frontier. Two design principles
that do not require a stronger reader remain available. The first is
heterogeneous redundancy: cross-checking by readers with decorrelated
failure modes --- different model families, differently-instructed roles,
and human reviewers --- whose joint miss rate can be driven down even when
no individual reader outperforms the generator. The second is mechanical
anchoring: pushing load-bearing claims down to artifacts that can be
checked without judgment --- thresholds frozen before data are seen,
re-runnable seeded computations, content hashes --- thereby shrinking the
surface on which any reader's judgment is load-bearing. The audit protocol
of this paper (preregistration, cross-family adversarial review, bit-exact
recomputation) instantiates both principles. We offer it not as a
solution: the residual class of errors that appear plausible to every
available reader remains, by construction, unmeasured. It is a reduction
of exposed surface, not an elimination of it.

\textbf{What this adds up to.} A cheap static audit predicted where the
reward would leak, in all three families (\S\ref{sec:track},
\S\ref{sec:xfam}). The train-side exposure it points at is large and
classifier-free: leak-vs-clean rewarded-FP gaps of \ptrainshare\,pt
(family A), \gatecC\,pt (B), and \famCcthree\,pt (C). The channel's
contents predate training --- the base models already produce the wrong
outputs the leaky suite pays for --- so what RLVR does here is select,
not learn (\S\ref{sec:mech}). And through all of it, held-out damage
stays bounded below \Xmargin\,pt (\S\ref{sec:avg}). At 1--1.5B and
\nsteps\ steps, a weak suite is a measurement problem before it is a
capability problem, and the measurement problem is auditable before
training ever starts.

\section{Limitations}
\label{sec:limitations}

\textbf{Scope.} Everything here is measured at 1--1.5B parameters,
\nsteps\ GRPO steps, and MBPP-length tasks. The bounded average and the
absence of measured growth are one-way statements about that box, not
clearances beyond it: reward-hacking pressure is reported to grow with
task length \citep{zhao2026specbench}, a regime our short tasks never
enter, and \citet{wang2026verification} argue that no fixed reward
survives capability growth. We state the continuation hypothesis in
falsifiable form: if exploitation of natural FPs grows, it should appear
as a rising FP-only-group share or a rising rewarded-FP frequency at
longer horizons or larger scale. A preregistered test at twice the
horizon came back negative --- a powered refutation in family B, an
underpowered non-detection in family C (\S\ref{sec:cont800}) --- so the
open axis is scale, and the same 250-task set means the task axis is
untested. All findings also live in one benchmark family
(MBPP/MBPP+, with HumanEval+ as the out-of-domain split); the HE+
transfer test was null, so every concentration statement is in-family
only.

\textbf{Inference limits.} The eval-side stratification is permanently
exploratory: the registered P3 was train-side, and the eval-side cut was
made after seeing the data (\S\ref{sec:strat}). Its specification-search
correction prices one axis of analyst freedom, so the adjusted values are
lower bounds; its magnitude is concentrated (the top five carriers hold
\pconc\%); and the high-leakiness stratum is a measurement floor
(\highleakbasezero\ of 20 tasks have base rate zero). No causal sentence
attaches to family-A data: separating ``the leak damages these tasks''
from ``these tasks attract FPs for other reasons'' needs per-sample eval
dumps and retained checkpoints, and the family-A grid kept neither ---
the family-B symmetry test modulates interpretation only
(\S\ref{sec:xfam}). Five seeds is the deliberate inference unit: exact
tests floor at $p=\pexactfloor$, and true gaps between roughly 0.2 and
\Xmargin\,pt are undetectable (\S\ref{sec:avg}). Step-0 twin structure is
family-specific, and the archived rollout schema is asymmetric across
arms (\S\ref{sec:honesty}).

\textbf{Instrument limits.} ``Hardened'' does not mean perfect: the audit
found plus-suite defects --- false negatives on correct code --- so the
hardened reward has its own, smaller error floor, quantified rather than
zero (\S\ref{sec:vw}). Audit depth is uneven across families: family A
took four signed predicate generations to reach \vwsharevfive\%, while
family C's \famCvwsharevfive\% is the least audited and includes
\famCsyntaxfail\% syntax-invalid extracted code among its failures; the
three-family share table must be read with that asymmetry, never as
``family C is harmed most.'' Families ran on homogeneous-but-different
hardware (A and B on RTX 3090s, C on 4090s), so we make no cross-family
numeric comparisons, only directional ones. Finally, the sub-behavioral
sharpening test bounds effects only above
\sharpfloorlo--\sharpfloorhi\ nats per token, only on families B and C
(family A has no retained final checkpoints); parameter-level movement
below that floor remains unprobed.

\textbf{Suite operationalization.} As preregistered and implemented,
both the hardened reward and \emph{all} held-out scoring run the
MBPP+/HumanEval+ extra tests only --- not the official EvalPlus union
scoring, which requires passing base and extra together. We flag this
prominently
because ``MBPP+'' colloquially denotes that union-scored metric. The two
operationalizations do not contain one another (MBPP+:
\disjointmbpp/\mbpptotal\ tasks fully disjoint; HumanEval+:
\disjointhe/\hetotal). The training-side consequence: across all
\relabelruns\ hardened runs, \relabelrewardn\ of \relabelrewardN\
positively rewarded rollouts (\relabelrewardpct\%;
\relabelmainn/\relabelmainN\ = \relabelmainpct\% in the main grid) pass
the extra tests while failing the base tests --- rewards union scoring
would deny. These are reward events, not distinct programs
(\relabelsig\ unique task-code signatures on \relabeltasks\ tasks). The
held-out consequence: every reported pass rate is extra-tests-only. We
re-scored the held-out set for families B and C under the stricter
official base$\wedge$plus union (family A's per-sample dumps were not
retained). Union scoring lowers both arms' pass rates near-symmetrically
(\bthreeshiftlo--\bthreeshifthi\,pt per arm, arm-to-arm difference at
most \bthreearmdiff\,pt), so every
family$\times$benchmark$\times$horizon cell stays on the same side of
the preregistered \Xmargin\,pt margin as under extra-only scoring,
leaving the pooled non-inferiority conclusion unchanged; the analysis
was preregistered before unblinding with a commitment to report either
way. The only cells outside the margin --- family C's two MBPP+ cells
--- sit outside it under \emph{both} scorings, reflecting family C's
wider intervals (\S\ref{sec:xfam}) rather than the scoring choice. The
false-positive
definition itself (base-pass $\wedge$ extra-fail) is unaffected by any
of this. The complete erratum file and the audit records behind this
disclosure will be released together with the code repository (see the
Reproducibility Statement).

\textbf{Meta-audit scope.} The meta-audit of \S\ref{sec:meta} carries its
own limits, stated in full in App.~\ref{app:meta} and summarized here. It
has two frontier judge subjects (plus one witness reader), MBPP only, one
frozen instrument; it supports no ``frontier models'' generalization. Its
self-assessment gap is measured across different item pools, so it is a
difference between settings and no causal reading attaches. Its deconfounded
same-author contrast returned UNRESOLVED --- a same-author deficit is neither
shown nor excluded. Its within-task contrasts cover zero, so it makes no
per-sample detection claim. The GPT-5.6 witness is a single blind session
that carries zero confirmatory weight by design and is biased upward by known
7/10 repository exposure; nothing in it is a bound. ``Supervision collapses
with capability'' is not established here as a law; we use it only as
motivation framed as an observation on this instrument, and the tested
generator axis is silent, in both directions, about generators not tested.
The two subjects are frontier-lab models but below the strongest public tier
at the time of writing, and MiMo-v2.5-Pro's placement is lower-confidence
(no government-grade third-party evaluation); the full generator-tier scope
note is in App.~\ref{app:meta}.

\section*{Reproducibility Statement}
\label{sec:repro}

The following are released with this paper in the code repository,
\url{https://github.com/toffee-desuwa/rlvr-leaky-suite} (items not yet
mirrored there are available to reviewers on request): the frozen task split (\ntrain\ train /
\nevalmbpp\ MBPP eval / \nevalhe\ HumanEval+; split RNG seed
\splitseedv); the pinned suite versions (evalplus 0.3.1) and both reward
paths; the full preregistration chain (v1, amendments, v2, v3) with its
failure ledger, signed adjudication cards, and the marker files that gate
the analysis scripts (\S\ref{sec:prereg}); the static leakiness tables
for all three families; the rollout logs, checkpoint evaluations, and
classification outputs behind every reported number; the predicate
specifications (v2--v5) with per-task adjudication dossiers; and the
analysis and audit scripts, including the independent recomputation
scripts used to verify headline quantities. Every number in this paper is
bound to a named macro whose definition records its source file. Training
hyperparameters are in \S\ref{sec:arms}; the evaluation protocol
($k=\ktemp$ at temperature 0.8, extra-test-suite oracle, seed as inference
unit) is in \S\ref{sec:evalproto} and \S\ref{sec:discipline}; known
archived-data caveats (rollout schema asymmetry, family-specific step-0
twins) are documented in \S\ref{sec:honesty} and in the data release.
The exploratory meta-audit of \S\ref{sec:meta} is released with this paper: its
five frozen preregistrations (the dual-role, MiMo (with Amendment~1),
cross-judging, attribution, and frontier-judge documents), the five signed
chair rulings and the headline annex, the per-cell row files and the three
frozen numbers registers, and the analysis scripts that reproduce every
meta-audit quantity from those rows. Per a signed scope amendment
(\texttt{AMENDMENT\_v1\_scope\_20260712}; pointer in all three registers) the
program enters this arXiv~v1 submission as exploratory material outside every
confirmatory claim, with all frozen sentences, bans, and downgrades in force.

\section*{LLM Usage Disclosure}
\label{sec:llm}

Large language models played a significant role in this work, and we
describe it plainly: they implemented and ran the training and analysis
code, orchestrated the preregistered analyses, drafted the paper's prose,
and ran internal adversarial reviews of claims and numbers; we name no
authoring tools. Separately, the exploratory meta-audit of \S\ref{sec:meta}
studies specific large language models as its experimental \emph{subjects}
(named there because they are the objects of measurement, not the tools that
produced this work), which is a distinct role from the vendor-neutral
authoring assistance described here. Assistants
from a second, independent model family provided blind replications and
cross-checks of several analyses, including a blind two-phase
reconciliation of the growth question. The study design was developed in
collaboration between the human author and these assistants; the human
author held final authority over every design decision, froze and signed
every preregistration document and amendment, made every per-task
adjudication and claim-strength ruling in the audit (each recorded in a
signed, versioned card in the released archive), and read and approved
all prose. Two mechanical safeguards backed this division of labor: no
verified-wrong aggregate could be computed until the human-signed
specification marker existed (\S\ref{sec:vwspec}), and headline
quantities were independently recomputed from raw logs before citation
--- in several cases by a second, isolated model instance given only the
raw data --- with numbers that failed recomputation removed rather than
cited. All remaining errors are the human author's responsibility.

\bibliography{refs}
\bibliographystyle{iclr2026_conference}

\clearpage
\appendix

\section{Gate ledger}
\label{app:gates}
The two failed launch gates of \S\ref{sec:prereg}, in full. Gate b (blind
hand-audit, 24 samples): VOID --- \gatebvoid\ verdicts hinged on
suite-side properties; its diagnosis established \scorerdiv\
scorer-vs-evalplus divergences on the same 24 samples. Gate b$'$
(evidence-grounded, one-shot, 20 verified-wrong samples): FAIL ---
\gatebprime\ overturned on evidence, with a post-mortem best estimate of
7/20 including one missed gray flip. The pass probability of gate b$'$
was computable from the signed document's own numbers at signing time
($\approx$1\%), which is what the gate-power rule of \S\ref{sec:prereg}
now prevents. Quoting rule, preserved from the registration: these
overturn counts come from a pool of motivated origin (high-FP rollouts),
five distinct tasks; task-level extrapolation only.

\section{Verified-wrong audit: taxonomy and worked examples}
\label{app:vw}
Table~\ref{tab:app-vw-perseed} gives the per-seed verified-wrong counts
for all three families; the family-A seed recovery method (rollout-level
re-join; zero ambiguous records, pooled total reproduced exactly) is
stated in its caption and referenced from \S\ref{sec:honesty}.

Below, one worked example per non-model cause, from the signed
adjudication dossiers. Each came out of auditing the \emph{hardened}
suite as skeptically as the leaky one. Throughout this appendix ``the
hardened suite'' denotes the extra-test reward suite --- the MBPP+
``plus'' tests only (extra-tests-only), not the base$\cup$plus union ---
and the counts below are per-task audited dispositions of those extra
tests.

\textbf{Plus-suite false negative --- Mbpp/806
(\texttt{max\_run\_uppercase}).} The stored expected values are produced
by running the dataset's canonical solution, and that canonical is
buggy: it overwrites its running maximum (\texttt{res = cnt}) instead of
taking a max, so 48 of 102 plus tests demand answers a correct
implementation does not produce (for \texttt{'Aaa'} it demands 0; the
correct answer is 1). A fully correct solution is forced to fail the
hardened suite. The signed predicate amnesties exactly those failures;
five records that also fail on other, in-spec tests remain
verified-wrong.

\textbf{Contract-invalid inputs --- Mbpp/267 (\texttt{square\_Sum}).}
The task's own contract demands \texttt{isinstance(n, int)}, but three
plus inputs are floats (\texttt{1000000.0}, \texttt{1e7}, \texttt{1e8}).
Any integer-domain solution crashes with a \texttt{TypeError} there; the
canonical survives only through silent float arithmetic, and its
division-based formula returns a float that the effective oracle then
prefers over the mathematically exact integer. All 97 of this task's
records were adjudicated suite-side; none is a genuine FP.

\textbf{Canonical convention --- Mbpp/639 (\texttt{sample\_nam}).} The
spec says to remove names that start with a lowercase letter. The
canonical additionally requires \texttt{el[1:].islower()}, a clause with
no basis in the spec sentence, so spec-correct code fails 38 of 111 plus
tests. The signed predicate (the broad reading-divergence rule)
amnesties failures where the canonical disagrees with a defensible
reading of the spec, covering this task's 73 records.

\textbf{Spec-silent boundary --- Mbpp/771 (\texttt{check\_expression}).}
The plus suite contains an empty-string test whose expected value the
spec never determines (is the empty expression balanced?); the canonical
answers by fiat. For 60 of 81 records that test is the sole failure, and
the signed ruling classifies them as spec-ambiguous rather than wrong;
the remaining 21 fail on other grounds and stay verified-wrong. This
task is also the largest persistent channel of \S\ref{sec:mech}.

The resource axis has its own exemplar, Mbpp/781: functionally correct
but slow solutions collect leaky reward and are killed by the hardened
suite's large inputs --- a failure axis orthogonal to correctness, which
is why Table~\ref{tab:vw} co-reports the 781-excluded sensitivity row.

\section{Estimator surfaces}
\label{app:surfaces}
Table~\ref{tab:app-p1-estimators} reports the full six-estimator CI
surface behind the \S\ref{sec:avg} bound. Table~\ref{tab:app-p3-strata}
reports the stratum-by-checkpoint H$-$L gaps behind \S\ref{sec:strat};
its caption carries the mandatory descriptive-aggregate rider for the
apparent deepening, and \texttt{null16} is included as the negative
control. All held-out pass rates in these tables --- and in
Table~\ref{tab:app-specsearch} --- are scored extra-tests-only (the MBPP+
``plus'' tests, not the base$\cup$plus union); family A's per-sample eval
dumps were not retained, so its held-out figures cannot be re-scored under
the union.

\section{Specification-search audit}
\label{app:specsearch}
Table~\ref{tab:app-specsearch} lists all \ssnvalid\ valid specifications
with their gaps, $t$ statistics, and ranks; the three anchor strata are
marked. The audit prices one axis of analyst freedom (the stratum
boundary), so its family-adjusted $p$-values are lower bounds on the true
correction (\S\ref{sec:strat}).

\section{Mechanism receipts}
\label{app:mech}
Table~\ref{tab:app-killtestA} gives the base-model control per persistent
channel, with all three denominator conventions in the caption.
Table~\ref{tab:app-r3} gives both readings of the family-B symmetry test,
including the clean-stratum control and per-seed sign patterns. The L4
group-level denominators are in the \S\ref{sec:mech} footnote.
Table~\ref{tab:app-cont800} gives the 800-step continuation receipts
(\S\ref{sec:cont800}).

\paragraph{Union-sensitivity operationalization (B3).} The union and
per-arm shift quantities in \S\ref{sec:limitations} are task-level
macro-averaged pass rates inheriting the paper's evaluation path
(\texttt{oracle\_eval.py}); a validity gate reproduced the extra-only
rates from the raw evaluation dumps bit-exactly (worst absolute
difference $0$). The non-inferiority read is the seed-paired H$-$L
one-sided 95\% upper bound (the CI90 upper end) against the
\Xmargin\,pt margin, inheriting the preregistration's decision rule
(PREREGISTRATION.md:49--51). The released four-cell pass/fail
decompositions are row-level micro scores --- a different granularity
from the macro rates --- and are descriptive only, entering no verdict.

\input{appendix_tables}


\section{Meta-audit: per-study receipts, frozen readings, and caveats}
\label{app:meta}

This appendix backs \S\ref{sec:meta}. Everything here is exploratory
\regE\ and descriptive; none of it entered the paper's preregistration and
none of it is required for any result in \S\ref{sec:results}. The program
has five audited studies, each with its own frozen preregistration and a
signed chair ruling: Meta-A (DeepSeek-V4-Pro dual role), Meta-A2
(MiMo-v2.5-Pro dual role), Meta-A3 (cross-judging deconfound), the GPT-5.6
witness lane inside Meta-A3, and Meta-A4 (explicit attribution). Two
subjects plus one witness reader, MBPP, one frozen instrument: the program
licenses nothing about frontier models in general, and no per-sample
detection claim for any cell (within-task contrasts cover 0). All intervals
are task-cluster bootstrap 95\% CIs ($B{=}10{,}000$); CIs are plain
percentile (BCa checked concordant). Two frozen rounding twins are preserved
deliberately and are not harmonized: the DeepSeek self-versus-anchor gap
prints as $-.276$ in the difference-CI table and $-.277$ in the licensed
synthesis sentences, and the MiMo difference-CI upper endpoint prints as
$-.058$ in the tables and $-.057$ in the MiMo licensed sentence; both members
of each pair are frozen text. Seeds were fixed at analysis time and
disclosed (primary bootstrap seed 20260711 for both subjects; difference CIs
20260715; anchor 20260712; Meta-A3/A4 joint-frame 20260713 with five
concordant sensitivity seeds; per-cell $D$ 20260717); the 378-task frame is
inert beyond the frozen set. Committed artifacts are left as-run (no silent
rewrites); one stale-provenance string in the MiMo metrics file is corrected
by note only, with the frozen numbers unaffected.

\subsection{Meta-A / Meta-A2: the dual-role gap (both subjects)}

\begin{table}[t]
\centering
\footnotesize
\setlength{\tabcolsep}{4pt}
\caption{Cross-subject descriptive readouts (preregistered readout,
Meta-A2). Descriptive only; two subjects license no generalization. The
self-assessment AUC and the hetero-judge anchor are on different item pools
(a judge's own FPs vs the 1.5B policy's FPs), so ``Self $-$ anchor'' is a
difference between settings, not a self-vs-hetero effect estimate.}
\label{tab:meta-dual}
\begin{tabular}{lcc}
\toprule
Readout & DeepSeek-V4-Pro & MiMo-v2.5-Pro \\
\midrule
Generations ($378$ MBPP $\times\,k{=}4$)                 & $1512$ & $1512$ \\
base-pass (rate)$^{\dagger}$                              & $1442$ ($.954$) & $1285$ ($.850$) \\
own-FP (rate among base-passers)                          & $217$ ($.150$) & $175$ ($.136$) \\
Self-assessment AUC [95\% CI]                             & $\maSelfDS$ $\maSelfDSci$ & $\maSelfMM$ $\maSelfMMci$ \\
Hetero-judge anchor (1.5B-FP pool, pooled)               & $\maAnchDS$ $\maAnchDSci$ & $\maAnchMM$ $\maAnchMMci$ \\
Self $-$ anchor, two-sample cluster boot$^{\ddagger}$    & $-.276$ $[-.352,\,-.200]$ & $-.148$ $\maDiffMMcitab$ \\
Spearman own-FP-rate vs static leakiness (\regE)         & $.605$ ($n{=}338$) & $.631$ ($n{=}334$) \\
FP-bearing tasks: all-FP / mixed                         & $47/18$ (of $65$) & $42/24$ (of $66$) \\
FP rows w/o within-task genuine comparator               & $184/217$ ($84.8\%$) & $136/175$ ($77.7\%$) \\
own-FP flagged at $p_{\mathrm{pass}}{<}0.5$ (diag.)$^{\S}$& $64/217$ ($29.5\%$) & $17/175$ ($9.7\%$) \\
genuine flagged at $p_{\mathrm{pass}}{<}0.5$ (diag.)$^{\S}$& $175/1225$ ($14.3\%$) & $37/1110$ ($3.3\%$) \\
Program cost                                             & about RMB\,20 & RMB\,16.44 \\
\bottomrule
\end{tabular}
\end{table}

$^{\dagger}$ \emph{Intent-equalized budget (AMENDMENT1).} The MiMo
generation stage ran at \texttt{max\_tokens} 4096 vs DeepSeek's 2048,
because MiMo's hidden-reasoning tokens count against its completion cap; at
2048, 32.5\% (13/40) of MiMo pilot generations were empty vs DeepSeek's
2.8\%, and the 4096 pilot gate read 0.0\% empty ($n{=}40$). Base-pass rates
therefore compare equalized-by-intent, not nominally identical, instruments.

$^{\ddagger}$ The earlier ``non-overlapping CIs, margin .020'' framing is
retired for the MiMo arm: its raw margin is convention-sensitive (compresses
to about .007 under BCa; never flips sign) and is not load-bearing. Both
cells report the two-sample task-cluster bootstrap on the AUC difference
(percentile, $B{=}10{,}000$, seed 20260715). The DeepSeek cell may keep its
conservative ``non-overlap, margin .168'' slack; the MiMo cell may not. The
item-pool confound is untouched by this instrument: each difference compares
two settings, not a self-vs-hetero effect. \emph{Frozen-text convention
note (T7): the MiMo upper bound is $-.058$ in this table row and $-.057$ in
the licensed sentence below; both are frozen and each is correct in its own
context.}

$^{\S}$ Diagnostic, non-preregistered, feeds nothing, firewalled;
operationalization $p_{\mathrm{pass}}{<}0.5$. The categorical verdict text
diverges on a few rows (MiMo verdict-``fail'' on own-FPs $26/175{=}14.9\%$;
on genuine $56/1110{=}5.0\%$). MiMo's low own-FP flag rate is judge house
style, not self-leniency: its stage-0 hetero profile was already
conservative (TPR .545, TNR .895) and it flags only 3.3\% of its own
genuine rows. Concealment / self-leniency framings are banned.

\paragraph{Frozen licensed reading --- DeepSeek (canonical long form).}
``When judging code drawn from its own output distribution,
DeepSeek-V4-Pro's self-assessed pass probability separates its own false
positives from genuine solutions \textbf{only weakly} (AUC .617,
task-cluster-bootstrap 95\% CI $[.555, .679]$ --- \textbf{above chance but
far below} the .894 $[.847, .936]$ the frozen hetero-judge protocol recorded
on a \textbf{different pool of FPs from a 1.5B policy}); \textbf{because the
two item pools differ}, and because \textbf{most of the signal is
between-task} ($47/65$ FP-bearing tasks are all-FP, and per-task FP rate
correlates with static leakiness, exploratory $\rho{=}.60$), the $-.277$ gap
is a \textbf{descriptive difference between settings}, and the residual
discrimination is \textbf{consistent with task-level leakiness recognition
rather than per-sample detection of its own errors}.''

\paragraph{Frozen licensed reading --- MiMo (canonical long form).}
``On its own-distribution errors, MiMo-v2.5-Pro's self-assessment
discriminates false positives from genuine passes \textbf{weakly but above
chance} (AUC .639, task-cluster-bootstrap 95\% CI $[.590, .690]$),
\textbf{descriptively below} its own hetero-judge anchor of .786
$[.710, .860]$ measured on \textbf{a different item pool} (the 1.5B policy's
FPs); a two-sample task-cluster bootstrap on the AUC difference gives
$-.148$, 95\% CI $[-.236, -.057]$, excluding zero --- but \textbf{because the
item pools differ, this is a difference between two settings, not an estimate
of a self-vs-hetero effect}; most of the residual signal is between-task
($42/66$ FP-bearing tasks are all-FP; $136$ of $175$ FP rows have no
within-task genuine comparator).''

\paragraph{Frozen two-subject convergence.}
``The between-task dominance and the FP-rate--leakiness correlation replicate
across both subjects ($\rho{=}.605$ and $.631$, \textbf{exploratory in both
preregs}), making task-level leakiness recognition \textbf{the parsimonious
account of the residual self-assessment signal in this pipeline}; this
two-subject convergence \textbf{is exploratory, is specific to MBPP under this
frozen instrument}, and \textbf{is flagged as the confirmatory target for the
Meta-A3 prereg rather than as a finding about models in general}.''
(Meta-A3's deconfounded cross-judging subsequently returned UNRESOLVED on its
confirmatory contrast --- see below; it neither confirmed the leakiness
account nor excluded a same-author effect.)

\subsection{Meta-A3: cross-judging (the deconfound)}

The four cells score $1-p_{\mathrm{pass}}$ against the own-FP label on
identical row sets per pool (DS pool $1442$ rows, MiMo pool $1285$ rows).
Integrity: all four bijective, parse-fail $0$; the off-diagonal cross cells
are the diagonal cells' stage-3 outputs re-scored (hash-verified). The
off-diagonal cells carry no individual CI in the frozen readouts and are
point estimates only.

\begin{table}[t]
\centering
\scriptsize
\setlength{\tabcolsep}{4pt}
\caption{Meta-A3 four cells (AUC) and frozen contrasts (joint 378-task-frame
bootstrap, seed 20260713). The confirmatory $\Delta_{\mathrm{DS}}$ returned
\textsc{unresolved}. All within-task contrasts cover 0. The $\Sigma$ row
displays the frozen value; it is narrated only via the frozen $\Sigma$
paragraph below.}
\label{tab:meta-a3}
\begin{tabular}{lcc}
\toprule
 & DS pool ($1442$) & MiMo pool ($1285$) \\
\midrule
Author (self) & $A_{DD}{=}.6172$ & $A_{MM}{=}.6388$ \\
Cross judge   & $A_{MD}{=}.5918$ & $A_{DM}{=}.5908$ \\
\midrule
\multicolumn{3}{l}{\emph{Contrasts}}\\
Confirmatory $\Delta_{\mathrm{DS}}{=}A_{MD}-A_{DD}$ & \multicolumn{2}{c}{$-.025$ $[-.090,\,+.040]$ --- covers 0 (\textsc{unresolved})} \\
Secondary $\Delta_{\mathrm{MiMo}}{=}A_{DM}-A_{MM}$  & \multicolumn{2}{c}{$-.048$ $[-.109,\,+.012]$ --- covers 0} \\
Secondary $\Sigma{=}\Delta_{\mathrm{DS}}+\Delta_{\mathrm{MiMo}}$ & \multicolumn{2}{c}{$-.073$ $[-.134,\,-.013]$ (frozen $\Sigma$ paragraph below)} \\
Within-task $W$ (self / hetero; contrast) & $.685/.574$; $-.111\,[-.287,+.074]$ & $.431/.559$; $+.128\,[-.080,+.333]$ \\
Falsification $I$ (400-row ref.) & \multicolumn{2}{c}{$-.005$ $[-.097,\,+.086]$ --- covers 0} \\
\bottomrule
\end{tabular}
\end{table}

\paragraph{Frozen synthesis (verbatim).}
``The $-.277/-.148$ Meta-A/A2 gaps therefore show \textbf{no evidence of} a
same-author deficit in this deconfounded test --- on identical items there is
\textbf{no evidence at this resolution} that either author ranks its own rows
worse than the cross reader (confirmatory $\Delta_{\mathrm{DS}}$ $-.025$
$[-.090, +.040]$, covers 0 $\rightarrow$ frozen ruling \textbf{UNRESOLVED};
$\Delta_{\mathrm{MiMo}}$ $-.048$ $[-.109, +.012]$, covers 0) --- \textbf{and a
same-author deficit is not excluded}: a penalty cancelled by an opposing
reader-capability difference remains an un-excluded explanation. The
parsimonious account \textbf{consistent with} these data is item-pool
composition, not same-author status: all four cells sit in a narrow
.59--.64 band, far below the same judges' hetero anchors on the 1.5B pool
($.786/.894$); own-distribution frontier FPs were weakly separable \textbf{at
the pool level} for both frontier judges tested, author or cross, while the
within-task contrasts cover 0, so \textbf{no per-sample detection reading is
licensed}.''

\paragraph{Frozen register note (verbatim).}
``Register note: this is the \textbf{pre-registered first deconfounded test}
of the same-author question; its confirmatory contrast returned
\textbf{UNRESOLVED} (neither a same-author deficit nor its absence is
established). Both directions were declared publishable at freeze.''
The non-equivalence phrasing is fixed: ``the cross judge did not detectably
rank the rows differently from the author at this resolution (both $\Delta$
cover 0; point estimates slightly below the author).'' And the precedent that
governs every sentence here: ``\,`no evidence of X' reports a failed detection
at a stated resolution; `not attributable to X' asserts X has been ruled out
as a cause. Meta-A3 earned the first and only the first.'' ``Parsimonious
account'' language survives only in ``consistent with''-modality with the
alternative kept alive in the same breath.

\paragraph{Frozen $\Sigma$ paragraph (verbatim; used wherever $\Sigma$ is
narrated).}
``$\Sigma$, the raw $2{\times}2$ interaction contrast (algebraically
$\Delta_{\mathrm{DS}} + \Delta_{\mathrm{MiMo}}$), is $-.073$, 95\% CI
$[-.134, -.013]$, excluding 0 on the negative side (joint 378-task-frame
bootstrap per frozen \S5; empirical two-sided $p \approx .018$): summed over
the two pools, the cross-judge cells rank lower than the author cells.
\textbf{Per the frozen table this is reported symmetrically with no
additivity-model conclusion in either direction} --- in the saturated
$2{\times}2$, $\Sigma$ is the single interaction degree of freedom, and
\textbf{this design cannot distinguish same-author familiarity from any other
judge$\times$pool interaction} (e.g., difficulty- or style-relative-to-judge).
Each individual within-pool contrast covers 0, so \textbf{no per-pool sentence
is licensed; both point estimates are negative, stated as point estimates
only}. The CI's tightness comes from the mandated joint draw, which cancels
the shared judge main effect (the mis-specified independent-pool variant is
wider and covers 0 --- \textbf{recorded as an audit fact, not an alternative
inference}). \textbf{Nothing confirmatory rests on $\Sigma$, and its direction
is the opposite of a same-author deficit}.'' (Coupling disclosure: the
replicate correlation between the two $\Delta$s is $-.533$; joint SD $.031$ vs
independent $.045$; the independent-pool variant $[-.162, +.016]$ covers 0 and
is mis-specified, recorded as an audit fact only.)

\paragraph{Frozen big-picture ammunition (verbatim).}
``Crossing the two frontier judges over the two locked frontier pools puts all
four judge$\times$pool cells in a narrow band (AUC .592--.639) regardless of
which judge reads which pool, while the same two judges score .894 and .786 on
the 1.5B policy's FPs (\textbf{different item pool and protocol stage;
descriptive comparison} --- the audited two-sample difference CIs exist for the
diagonal cells, $-.276$ $[-.352, -.200]$ and $-.148$ $[-.236, -.058]$, both
excluding 0, and the cross cells sit lower still in point estimate).
Descriptively, \textbf{weak pool-level separability tracks the item pool, not
the reader and not authorship}: every frontier reading of own-distribution
frontier FPs tested here --- DS-on-DS, MiMo-on-DS, MiMo-on-MiMo, DS-on-MiMo ---
lands in the same weak band. This is a two-judge, two-pool observation on MBPP
under this instrument; \textbf{it licenses nothing about frontier models in
general and no per-sample detection claim} (both within-task contrasts cover
0).''

\subsection{GPT-5.6 witness lane (single blind session)}

\paragraph{Protocol and claim-ladder.} The witness lane ran at FULL tier,
$35/35$ batches, $2680$ rows ($={}2727-47$ exposed-six rows), $0$ integrity
problems; the chair reproduced every number to $<10^{-12}$ including both CIs,
and the single abstain-$0.5$ row (J002720) is \S7g-conformant. The lane was
built to carry \emph{zero confirmatory weight}, a design-level ceiling no
wording can unlock, under a frozen claim-ladder: (i) the only GPT interval is
a session-conditional, task-clustered percentile-bootstrap CI on the own-pool
separability point $C_p{=}A_{Gp}-0.5$, and every sentence on it is prefixed
``in this single blind session'' and is not a property of GPT-5.6 as a judge;
(ii) $\Gamma_p{=}A_{Gp}-A_{pp}$ is point estimate and sign only (no CI, no
``excludes 0''); (iii) the within-task contrast is descriptive direction only;
(iv) no ``$\approx$'', ``cannot separate'', equivalence, or ceiling framing;
(v) known contamination (7/10 exposure) plus within-run accumulation bias
$C_p$ \emph{upward}, stated wherever $C_p$ is quoted; (vi) the
single-session-vs-$N$-fresh trade is statistically worse and is acceptable
only because the lane is never confirmatory.

\begin{table}[t]
\centering
\footnotesize
\setlength{\tabcolsep}{5pt}
\caption{GPT-5.6 witness ladder (frozen \S6--7). Session-conditional CIs; upward
bias direction stated (7/10 exposure + within-run accumulation); never
confirmatory. $\Gamma_p$ is point and sign only. Within-task $W$ is
descriptive direction only.}
\label{tab:meta-witness}
\begin{tabular}{lcc}
\toprule
Readout & DS pool ($1418$; $217$ FP) & MiMo pool ($1262$; $175$ FP) \\
\midrule
$A_G$ (row-level AUC) & $.756$ $[.698,\,.812]$ & $.744$ $[.687,\,.803]$ \\
$\Gamma_p{=}A_G-A_{pp}$ (point$+$sign) & $+.136$ (diag $.621$) & $+.104$ (diag $.640$) \\
Length baseline (same rows) & $.655$ & $.674$ \\
Docstring baseline ($A_{GM}$ guard) & $.500$ & $.561$ \\
Within-task $W$ (descr.\ direction) & $.775$ ($18$ mixed) & $.635$ ($24$) \\
Per-file AUC dispersion & mean $.695$, sd $.088$ & mean $.621$, sd $.140$ \\
\quad order slope$^{\star}$ & $-.003$/file & $+.012$/file (flagged) \\
GPT-5.6 on 1.5B-FP pool & \multicolumn{2}{c}{$.959$ $[.930,\,.981]$ (contamination disclosed)} \\
\bottomrule
\end{tabular}
\end{table}

$^{\star}$ \emph{Frozen R1 annotation (verbatim, both pools incl.\ DS
$-.003$):} ``slope computed over the mixed-class files only (batches 1--18;
batches 19--35 contain no FP rows, so the diagnostic does not see the back
half of the session), and file order is collinear with the
W$\rightarrow$ENRICHED tier boundary --- the flag cannot distinguish instrument
drift from row composition; no temporal-drift reading is licensed.''

\paragraph{Frozen witness reading (verbatim, canonical long form).}
``The highest-scoring reader tested here is GPT-5.6: AUC .959 $[.930, .981]$ on
the 1.5B policy's FPs in the preregistered fresh-session blind probe
(\textbf{contamination disclosed at freeze --- 7/10 repository exposure,
exposed-six excluded, residual familiarity not ruled out}; the
contamination-free floor is DeepSeek-V4-Pro's .894 $[.847, .936]$). \textbf{In
this single blind session}, the same reader lands at .756 $[.698, .812]$ on the
DS frontier pool and .744 $[.687, .803]$ on the MiMo frontier pool
(\textbf{session-conditional CIs; never confirmatory}; both known bias sources
--- the same 7/10 exposure plus within-run accumulation --- push these numbers
upward, in the judge's favor: \textbf{correcting the known biases could only
lower them, while single-instance variance remains unquantifiable in both
directions}). The .959-versus-.74/.76 contrast is a \textbf{descriptive
comparison across different item pools and different protocol stages}
($400$-row fresh-session multi-shard vs $2680$-row single-session batch),
\textbf{not a same-instrument measurement}. With that grading, the three-reader
picture is descriptively ordered: all three readers tested score .786--.959 on
the 1.5B pool; the two frontier subjects reading own-distribution frontier FPs
sit in a \textbf{weak band (.592--.639, above chance}; the audited two-sample
differences from their own 1.5B anchors are $-.276$ $[-.352, -.200]$ and
$-.148$ $[-.236, -.058]$); and the highest-scoring tested reader improves on
that band ($\Gamma{=}+.136/+.104$, \textbf{point and sign only}; beats the
length and docstring baselines on both pools) yet stays far below every 1.5B
anchor. Three judges, three pools, MBPP, this instrument: \textbf{nothing here
licenses a claim about frontier models in general, about readers or generators
not tested, no per-sample detection claim} (within-task contrasts cover 0; the
GPT within-task direction is descriptive-only), \textbf{and no cause
attribution} --- item-pool composition, judge--pool interaction, and protocol
stage are not separable in this design.''

\paragraph{Frozen scope-update.} The block order is fixed: audited band
sentence $\rightarrow$ witness kicker $\rightarrow$ generator scope note.
``The pool-driven collapse from the 1.5B anchors (.786--.959) remains the
dominant structure, while the band's LEVEL is reader-dependent (.59--.64 API
judges vs .69--.81 session-conditional GPT interval).''

\subsection{Meta-A4: explicit attribution ($2{\times}2{\times}2$)}

The confirmatory family is $\{I(\mathrm{DS}), I(\mathrm{MiMo})\}$ with Holm:
DS tested at $.05/2{=}.025$ (rejected); MiMo at $.05$ (not rejected); the
degradation rule did not fire. The calibration downgrade rule (frozen \S1): a
calibration cell's $D_C$ CI excluding 0 downgrades all per-cell $D$ readouts
for that (judge, pool) to descriptive only, with drift\_rms co-reported as a
noise floor; $K$ and $I$ are unaffected, being attributed-cells-only
statistics in which the silent baseline cancels row-wise (chair-verified: delete
the baseline and $K$ and both $I$ reproduce to 10 dp).

\begin{table}[t]
\centering
\footnotesize
\setlength{\tabcolsep}{6pt}
\caption{Meta-A4 confirmatory interactions, claim-content $K$ contrasts, and the
calibration arm. The sign flip in the DS $K$ row IS the confirmatory
interaction; the full $2{\times}2$ is reported with no single-story
compression. $K$ contrasts are exploratory, per-comparison CI, outside the
family. Detection status is reported per judge; $I$/$K$/$D$ are never compared
across judges.}
\label{tab:meta-a4}
\begin{tabular}{lcc}
\toprule
 & DeepSeek judge & MiMo judge \\
\midrule
Confirmatory $I$ & $-.106$ $[-.178,\,-.035]$, $p{=}.0032$ (Holm) & $+.021$ $[-.020,\,+.064]$, $p{=}.32$ \\
$K$, own pool    & $-.043$ $[-.089,\,+.001]$ & $+.000$ $[-.034,\,+.033]$ \\
$K$, other pool  & $+.063$ $[+.013,\,+.118]$ (excl.\ 0) & $-.021$ $[-.050,\,+.007]$ \\
\midrule
\multicolumn{3}{l}{\emph{Calibration arm} ($D_C$; drift\_rms noise floors $.156$--$.322$)}\\
\multicolumn{3}{l}{\quad cal\_AMM $+.085$ $[+.021,\,+.154]$ \textbf{(excl.\ 0 $\rightarrow$ downgrades all (MiMo,MiMo) per-cell $D$)}} \\
\multicolumn{3}{l}{\quad cal\_ADD $-.062$ $[-.174,\,+.050]$; cal\_ADM $-.011$; cal\_AMD $+.041$ $[-.005,\,+.094]$ (cover 0)} \\
\bottomrule
\end{tabular}
\end{table}

Per-cell $D$ (task-cluster CI, seed 20260717; all cover 0; the (MiMo,MiMo)
cells$^{\dagger}$ are descriptive only per the calibration downgrade):
$Y_{DD}\,{-}.029\,[-.079,+.019]$; $O_{DD}\,{+}.015\,[-.033,+.063]$;
$Y_{DM}\,{+}.025\,[-.033,+.081]$; $O_{DM}\,{-}.038\,[-.092,+.014]$;
$Y_{MM}\,{+}.015\,[-.027,+.058]^{\dagger}$;
$O_{MM}\,{+}.015\,[-.022,+.050]^{\dagger}$;
$Y_{MD}\,{-}.018\,[-.049,+.011]$; $O_{MD}\,{+}.002\,[-.026,+.029]$. Net FP
flips: $O_{DM}\,{+}8.6\%$ (DS hard-fails MiMo FPs when told ``another
model's''; flip-differential $+.0596$), $Y_{MD}\,{+}4.1\%$ ($+.0374$); all
abstain counts $0$; $|\Delta p|{>}.5$ share $.05$--$.12$ per cell. Provenance:
$12$ cells, $10{,}908$ attributed $+\,600$ calibration $+\,12$ smoke rows;
parse-fail $0$, abstain $0$; MiMo side RMB\,10.98/20 cap, DS side RMB\,46.6/55
cap; an $O_{DD}$ Windows-update interrupt ($1277/1442$) was resumed by dedupe.

\paragraph{Frozen DS licensed sentence (verbatim, canonical long form).}
``For the DeepSeek judge, the confirmatory interaction is $I(\mathrm{DS}){=}
-.106$, 95\% CI $[-.178, -.035]$, two-sided bootstrap $p{=}.0032$,
\textbf{Holm-surviving at the .025 threshold} (family $\{I(\mathrm{DS}),
I(\mathrm{MiMo})\}$): the response to the authorship-claim pair differs between
the configuration where the YOURS claim is true and the configuration where the
OTHER claim is true, in the \textbf{harsher-when-truly-own direction}. Within
its own pool the claim pair moves the judge toward relative harshness on its own
false positives ($K{=}-.043$ $[-.089, +.001]$); on the other author's pool the
same claim pair moves it toward relative leniency on false positives ($K{=}
+.063$ $[+.013, +.118]$, \textbf{exploratory, outside the family}); \textbf{the
sign flip between the true-claim and false-claim configurations IS the
confirmatory interaction, and the full $2{\times}2$ is reported with no
single-story compression}. This pattern is \textbf{consistent with
truth-sensitivity of the response through any channel (stylistic familiarity
explicitly among them) AND, inseparably, with a claim-content $\times$
pool-property interaction that is not truth-mediated} (pool difficulty or
FP-type composition modulating susceptibility to authorship framing);
\textbf{both readings are named wherever this result is quoted. NOT licensed:
recognition/memory/knowing; channel attribution; `deception'; unqualified `the
shift is truth-sensitive'.}''

\paragraph{Frozen disjunction (verbatim; ``inseparably'' is frozen text).}
``the response to the authorship-claim pair differs between the true-claim and
false-claim configurations, in the \textbf{harsher-when-truly-own} direction ---
consistent with truth-sensitivity via any channel (stylistic familiarity among
them) \textbf{AND, inseparably}, with a claim-content $\times$ pool-property
interaction (difficulty/FP-type susceptibility); both named. NOT licensed:
recognition/memory/knowing; channel attribution; `deception'; unqualified `the
shift is truth-sensitive'.''

\paragraph{Frozen footprint-constraint (verbatim; closing clause mandatory).}
``Audit quantification (descriptive; exploratory label): for the DS judge the
claim-content main effect is small and nearly pool-invariant ($+.016$ on the own
pool vs $+.022$ on the other pool; pool difference $-.006$, about 5\% of $|I|$),
and the two pools are near-matched in FP fraction (15.0\% vs 13.6\%).
Essentially all of $I(\mathrm{DS})$ lives in the FP-versus-genuine differential,
whose content effect inverts sign across pools ($-.021$ on own-pool FP rows vs
$+.076$ on other-pool FP rows) while the genuine-row effect is stable ($+.023$
vs $+.013$). A non-truth-mediated account (branch B) must therefore supply a
pool-property susceptibility modifier that concentrates almost entirely on
false-positive rows and inverts sign between two pools of near-matched FP
composition. \textbf{This constrains the SHAPE the second branch must take; it
does not rank the branches, and this design cannot}.''

\paragraph{Frozen MiMo null (verbatim, canonical long form).}
``For the MiMo judge: \textbf{no detected attribution effect of either kind at
this resolution} ($I(\mathrm{MiMo}){=}+.021$, 95\% CI $[-.020, +.064]$,
$p{=}.32$, not rejected at Holm .05; both $K$ contrasts cover 0:
$K(\mathrm{mimo,mimo}){=}+.000$ $[-.034, +.033]$, $K(\mathrm{mimo,ds}){=}-.021$
$[-.050, +.007]$). \textbf{This is a non-detection, not truth-invariance and not
equivalence}; the both-judges compounding row does not fire (the DS judge
detected). The (mimo, mimo) calibration cell drifted by itself (cal\_AMM
$D_C{=}+.085$ $[+.021, +.154]$, excluding 0), so \textbf{all (mimo, mimo)
per-cell $D$ readouts are descriptive only} (frozen \S1 rule; reported, not
silently applied). \textbf{$K$ and $I$ are unaffected: they are
attributed-cells-only statistics in which the silent baseline cancels row-wise}
--- verified in audit by recomputing every $K$ and both $I$ with the baseline
deleted, reproducing all six to ten decimal places. What the drift does bound is
resolution: row-level $\Delta p$ is unstable under silent re-judging at
temperature 0 (drift\_rms $.16$--$.32$ across calibration cells), and this
non-detection is scoped to the resolution that noise floor permits.''

\paragraph{Frozen firewall connective (verbatim; the ONLY licensed connective if
Meta-A3's $\Sigma$ and Meta-A4's $I(\mathrm{DS})$ are placed adjacent).}
``Two adjacent results from different designs: Meta-A3's exploratory $\Sigma$
(silent cross-judging; reported symmetrically, no additivity-model conclusion in
either direction) and Meta-A4's confirmatory $I(\mathrm{DS})$ (explicit
attribution claims; frozen two-branch disjunction). They are different
statistics --- a rank-AUC interaction contrast versus a claim-conditioned
pass-probability interaction --- under different manipulations (no claim text
versus claim text), and \textbf{neither bears on the other's interpretation.
Sign agreement across incommensurable statistics carries no evidential weight
and is not narrated as convergence. Meta-A3's confirmatory same-author contrast
remains UNRESOLVED; Meta-A4 does not address it.}''

\paragraph{Registered leads (future work only).} Three exploratory leads are
recorded; pursuit requires a new signed preregistration and none is a rescued
claim: L1 $K(\mathrm{ds,mimo}){=}+.063$; L2 the $O_{DM}$ flip $+8.6\%$
($+.0596$); L3 the cal\_AMM self-drift $+.085$.

\subsection{Generator-tier scope note (verbatim)}

``Scope note on the generator axis: the two subject models, while frontier-lab
models, are \textbf{not the strongest publicly available tier at the time of
writing} (by public positioning, e.g. GPT-5.6 and Claude Fable 5 sit above
them). The weak-readability band we measure therefore occurs at a generator tier
below the current top, and \textbf{the concern it raises does not rest on
generators having reached any ceiling. The tested generator range does not
extend to that top tier, and these data are silent --- in both directions ---
about readability of FPs from generators not tested}: this design has two
generator levels, and the 1.5B-versus-frontier difference itself \textbf{is not
attributable to generator capability as a cause} (pool composition, judge--pool
interaction, and protocol stage are not separable).'' Factual predicate
(verified 2026-07-12): the subjects rank about \#16--17 on the AA Coding Index
(7/11), roughly two steps below the strongest, an about-8-month lag per
CAISI/NIST (2026-05-01); MiMo-v2.5-Pro has no government-grade third-party
evaluation, so its placement carries lower confidence and is flagged as such.
The symmetric claim ``readability would recover at higher tiers'' is equally
banned.

\subsection{Caveat ledger (complete)}

Every caveat below binds the whole section and is stated once here so it is not
diluted by repetition.
\begin{itemize}\itemsep2pt
\item[E1.] \textbf{Publication scope.} This program is exploratory and carries
zero confirmatory weight for the paper's claims (\S\ref{sec:results}); by a
signed same-day amendment (\texttt{AMENDMENT\_v1\_scope\_20260712}, Toffee,
2026-07-12; pointer recorded in all three frozen registers) it enters the
2026-07-14 arXiv~v1 submission, superseding the earlier ``v2-register-only''
routing. Every frozen sentence, ban, and downgrade remains in full force: the
exploratory status, the \texttt{cal\_AMM} calibration downgrade, the witness
claim-ladder, the Meta-A3 UNRESOLVED status, and the generator-tier scope note.
\item[E2.] \textbf{Item-pool confound.} Self-AUC and hetero anchor are on
different item pools; every gap is a difference between settings, not a
self-vs-hetero effect estimate. No causal language.
\item[E3.] \textbf{$N{=}2$ subjects ($+1$ witness judge).} Two frontier subjects
on MBPP under one frozen instrument; enumeration only; no ``frontier models''
generalization anywhere.
\item[E4.] \textbf{Between-task structure.} Most residual self-assessment signal
is between-task (DS 84.8\% / MiMo 77.7\% of FP rows lack a within-task genuine
comparator); .617/.639 may substantially reflect task-leakiness recognition, not
per-sample error detection.
\item[E5.] \textbf{MiMo raw-CI margin not load-bearing.} The retired
``non-overlap margin .020'' compresses to about .007 under BCa (never flips); the
load is the difference CI $[-.236, -.057]$. The MiMo gap is not the DS gap's
evidential comfort (DS keeps its conservative .168 slack).
\item[E6.] \textbf{AMENDMENT1 budget.} MiMo \texttt{max\_tokens} 4096 vs DS 2048
(hidden-reasoning tokens count against the cap; 2048 $\rightarrow$ 32.5\% empty
vs DS 2.8\%; 4096 gate 0.0\%, $n{=}40$); base-pass rates compare
equalized-by-intent instruments.
\item[E7.] \textbf{Diagnostic firewall.} The $p_{\mathrm{pass}}{<}0.5$ flag
counts (DS 29.5\% / MiMo 9.7\% own-FP) are non-preregistered, feed nothing, are
descriptive; MiMo's low rate is judge house style (flags 3.3\% of own genuine),
not self-leniency; operationalization stated ($p_{\mathrm{pass}}{<}0.5$ 9.7\% vs
verdict-text 14.9\%).
\item[E8.] \textbf{Meta-A3 UNRESOLVED.} The confirmatory contrast returned
UNRESOLVED; a same-author penalty cancelled by an opposing reader-capability
difference is un-excluded. Only ``no evidence of X'', never ``not attributable to
X''.
\item[E9.] \textbf{$\Sigma$ symmetric.} $\Sigma{<}0$ excludes 0 but licenses no
additivity-model conclusion in either direction; the independent-pool variant
covers 0 (audit fact only); nothing confirmatory rests on $\Sigma$.
\item[E10.] \textbf{Within-task null.} Both API within-task contrasts cover 0
$\rightarrow$ no per-sample detection sentence for any cell; the GPT within-task
direction is descriptive-only.
\item[E11.] \textbf{Single blind GPT session.} Every witness number is scoped
``in this single blind session''; $C_p$ carries a session-conditional CI (not a
property of GPT as a judge); $\Gamma_p$ is point and sign only; no
equivalence/ceiling; never confirmatory (design-level).
\item[E12.] \textbf{Upward-bias direction.} Known contamination (7/10) plus
within-run accumulation bias every witness separability number upward; correcting
known biases could only lower them; single-instance variance is unquantifiable in
both directions, so no bound language.
\item[E13.] \textbf{Cross-pool/cross-protocol.} .959-vs-.74/.76 is a descriptive
comparison across different item pools and protocol stages ($400$-row
fresh-session multi-shard vs $2680$-row single-session batch), not a
same-instrument measurement.
\item[E14.] \textbf{R1 order-slope caveat.} The MiMo $+.012$/file (and DS
$-.003$/file) slope is on mixed-class files only (batches 1--18; 19--35 FP-free),
file order collinear with the W$\rightarrow$ENRICHED boundary; no temporal-drift
reading.
\item[E15.] \textbf{.959 anchor contamination.} The GPT 1.5B-pool .959 carries
7/10 exposure (exposed-six excluded, residual familiarity not ruled out); the
contamination-free floor is DS .894 $[.847, .936]$; the disclosure travels
wherever .959 is cited.
\item[E16.] \textbf{cal\_AMM downgrade.} cal\_AMM $D_C{=}+.085$ $[+.021, +.154]$
excludes 0 $\rightarrow$ all (MiMo,MiMo) per-cell $D$ are descriptive only;
drift\_rms $.156$--$.322$ co-reported; $K$ and $I$ unaffected (baseline-free by
row-wise cancellation, chair-verified). The downgrade is reported, not silently
applied.
\item[E17.] \textbf{Disjunction inseparability.} The $I(\mathrm{DS})$ branches
stay strictly co-equal (``inseparably'' is frozen); no ranking adjective; the
footprint-constraint sentence is the only vehicle for the audit quantification
and closes with ``does not rank the branches, and this design cannot''.
\item[E18.] \textbf{Cross-judge non-comparison.} $I$/$K$/$D$ are reported per
judge; cross-judge comparisons are banned (attribution sentences may parse
differently per vendor).
\item[E19.] \textbf{Meta-A4 $\leftrightarrow$ Meta-A3 firewall.} Co-citation only
via the frozen connective; synthesis banned; sign agreement across incommensurable
statistics carries no evidential weight; Meta-A4 does not address Meta-A3's
UNRESOLVED.
\item[E20.] \textbf{Generator-tier scope.} Subjects are frontier-lab models below
the strongest public tier at the time of writing; the weak band occurs below the
top tier; the concern does not rest on any generator ceiling; the data are
two-way silent above the tested tier; ``top-tier generators would be even less
readable'' and its mirror are banned in all modalities; MiMo placement is
lower-confidence (no government-grade third-party evaluation).
\item[E21.] \textbf{Registered leads.} $K(\mathrm{ds,mimo}){=}+.063$, $O_{DM}$
flip $+8.6\%$, cal\_AMM self-drift $+.085$ are exploratory leads; pursuit needs a
new signed prereg, never a rescued claim.
\item[E22.] \textbf{Method notes.} Plain percentile CIs (BCa concordant); seeds
fixed at analysis time and disclosed; the 378-task frame is inert beyond the
frozen set; committed artifacts left as-run (corrective notes only).
\end{itemize}

\end{document}

%% file: math_commands.tex
\usepackage{amsmath,amsfonts,bm}

\def\eqref#1{equation~\ref{#1}}

\def\1{\bm{1}}

\DeclareMathAlphabet{\mathsfit}{\encodingdefault}{\sfdefault}{m}{sl}
\SetMathAlphabet{\mathsfit}{bold}{\encodingdefault}{\sfdefault}{bx}{n}



%% file: numbers.tex
\newcommand{\Xmargin}{1.5}          
\newcommand{\nseeds}{5}             
\newcommand{\nsteps}{400}           
\newcommand{\ktemp}{8}              
\newcommand{\ntrain}{250}           
\newcommand{\nevalmbpp}{128}        
\newcommand{\nevalhe}{164}          
\newcommand{\stoprule}{10}          

\newcommand{\pgapupper}{0.75}       
\newcommand{\pgapmean}{0.20}        
\newcommand{\pgapci}{[-0.42,\,+0.74]} 
\newcommand{\fptrendci}{[-0.012,\,+0.026]} 

\newcommand{\rhotrackraw}{0.80}     
\newcommand{\rhotrackpartial}{0.79} 
\newcommand{\staticrawmass}{15.30}  
\newcommand{\staticexclmass}{13.71} 

\newcommand{\vwshare}{68.84}        
\newcommand{\vwsharevthree}{59.91}  
\newcommand{\vwsharevfive}{47.57}       
\newcommand{\vwcountvfive}{1342}        
\newcommand{\vwnvfive}{2821}            
\newcommand{\vwcodededupvfive}{48.77}   
\newcommand{\vwcodededupcount}{1307}    
\newcommand{\vwcodededupn}{2680}        
\newcommand{\vwtaskmeanvfive}{0.808}    
\newcommand{\vwtaskmeanci}{[0.749,\,0.862]} 
\newcommand{\vwanyvwvfive}{94.3}        
\newcommand{\vwunresexclvfive}{49.81}   
\newcommand{\vwforcedfalsevfive}{47.43} 
\newcommand{\vwsharevfiveexclA}{48.85}  
\newcommand{\vwclusterci}{[36.4,\,60.5]} 
\newcommand{\vwearlyvfive}{49.1}        
\newcommand{\vwlatevfive}{46.0}         
\newcommand{\famBvwsharevfive}{45.37}   
\newcommand{\famBvwcountvfive}{976}     
\newcommand{\famBseedci}{[43.20,\,47.52]} 
\newcommand{\famCvwsharevfive}{62.78}   
\newcommand{\famCvwcountvfive}{1366}    
\newcommand{\famCseedci}{[61.78,\,63.75]} 

\newcommand{\rewardgap}{8.37}       
\newcommand{\fpmassshare}{8.82}     
\newcommand{\npersistent}{13}       
\newcommand{\npersistentrobust}{12} 

\newcommand{\ptrainshare}{+43.8}    
\newcommand{\ptrainci}{[43.05,\,44.57]} 
\newcommand{\ptrainleak}{45.9}      
\newcommand{\ptrainclean}{2.1}      

\newcommand{\ptrainstrat}{+1.60}    
\newcommand{\pstratci}{[+0.37,\,+2.82]} 
\newcommand{\pstratseeds}{5/5}      
\newcommand{\pcleanstrat}{-0.03}    
\newcommand{\pdid}{+1.63}           
\newcommand{\pdidp}{0.067}          
\newcommand{\pfloorthirty}{+2.58}   
\newcommand{\pmidband}{+2.78}       
\newcommand{\phighleak}{+0.00}      
\newcommand{\pexactfloor}{0.031}    
\newcommand{\pconc}{60.0}           

\newcommand{\ssnvalid}{106}         
\newcommand{\ssNaFamily}{0.091}     
\newcommand{\ssNbFamily}{0.094}     
\newcommand{\ssNtwoFamily}{0.393}   
\newcommand{\ssrankT}{43}           
\newcommand{\ssrankGap}{83}         

\newcommand{\scorerdiv}{0}          
\newcommand{\gatebvoid}{8/24}       
\newcommand{\gatebprime}{6/20}      

\newcommand{\famBmodel}{deepseek-coder-1.3b-instruct} 
\newcommand{\gatebC}{0.8129}        
\newcommand{\gatebCci}{[0.7599,\,0.8558]} 
\newcommand{\gatecC}{+45.6}         
\newcommand{\gatecCci}{[43.7,\,47.6]} 
\newcommand{\famCctwo}{0.6998}      
\newcommand{\famCctwoci}{[0.620,\,0.767]} 
\newcommand{\famCcthree}{+34.4}     
\newcommand{\famCcthreeci}{[32.5,\,36.4]} 
\newcommand{\bcTaskDelta}{-0.216}   
\newcommand{\cfourB}{-1.389}        
\newcommand{\cfourBci}{[-2.774,\,-0.004]} 
\newcommand{\cfourBp}{0.094}        
\newcommand{\famCsyntaxfail}{9.3}   

\newcommand{\condplusA}{0.824}      
\newcommand{\condplusB}{0.816}      
\newcommand{\condplusC}{0.805}      
\newcommand{\uncondlpA}{0.4997}     
\newcommand{\uncondlpB}{0.3681}     
\newcommand{\uncondlpC}{0.3548}     
\newcommand{\uncondhA}{0.5042}      
\newcommand{\uncondhB}{0.3702}      
\newcommand{\uncondhC}{0.3626}      
\newcommand{\naiveinclA}{2694}      
\newcommand{\naiveinclB}{2046}      
\newcommand{\naiveinclC}{2082}      

\newcommand{\staticxfamlo}{0.69}    
\newcommand{\staticxfamhi}{0.75}    

\newcommand{\regC}{[C]}             
\newcommand{\regE}{[E]}             
\newcommand{\regI}{[I]}             

\newcommand{\pgapsurfacelo}{0.70}   
\newcommand{\pgapsurfacehi}{1.07}   
\newcommand{\mdehalfboot}{0.58}     
\newcommand{\mdehalft}{0.80}        
\newcommand{\fpquarterfirst}{14.5}  
\newcommand{\fpquarterlast}{15.2}   
\newcommand{\learnmbpprange}{2.3--2.9} 
\newcommand{\learnherange}{1.6--1.7}   
\newcommand{\rhotrackbootci}{[0.74,\,0.85]} 
\newcommand{\olsbetaleak}{.896}     
\newcommand{\olsbetadiff}{-.049}    
\newcommand{\ntrackedtasks}{216}    

\newcommand{\rhoearly}{.77}         
\newcommand{\rholate}{.76}          
\newcommand{\fponlyA}{11.9}         
\newcommand{\fponlyAci}{[11.4,\,12.3]} 
\newcommand{\fponlyB}{13.7}         
\newcommand{\fponlyC}{13.9}         
\newcommand{\lfourfp}{94.6}         
\newcommand{\lfourhonestleak}{14.9} 
\newcommand{\lfourhonestclean}{98.8}
\newcommand{\lfourdenomcompl}{1130} 
\newcommand{\lfourdenomleak}{295}   
\newcommand{\lfourdenomclean}{1683} 
\newcommand{\lfourcleannull}{97.7}  
\newcommand{\zerogradmixed}{76.8}   
\newcommand{\distinctlargest}{55/55}
\newcommand{\basectrlchannels}{11}  
\newcommand{\basectrlfam}{9}        
\newcommand{\basectrlpooled}{7}     
\newcommand{\basectrlguard}{10}     
\newcommand{\fpdeclineBci}{[-0.033,\,-0.003]} 
\newcommand{\sharplenconf}{24--29}  
\newcommand{\genslopelo}{-0.54}     
\newcommand{\genslopehi}{+0.04}     
\newcommand{\genworstub}{+0.25}     
\newcommand{\genmaxrise}{1}         
\newcommand{\genlevel}{7}           
\newcommand{\npredtasks}{32}        
\newcommand{\basepassriseall}{15/15} 
\newcommand{\sharpfloorlo}{0.001}   
\newcommand{\sharpfloorhi}{0.0024}  

\newcommand{\rthreedeltaexposed}{0.000} 
\newcommand{\rthreep}{.56}          
\newcommand{\rthreesens}{+0.017}    
\newcommand{\rthreesensci}{[-0.001,\,+0.035]} 
\newcommand{\cfourCp}{.18}          
\newcommand{\zeroleakybetter}{0}    
\newcommand{\highleakbasezero}{17}  
\newcommand{\ptrainnleak}{77}       
\newcommand{\ptrainnclean}{173}     
\newcommand{\splitseedv}{20260702}  
\newcommand{\nevalpaired}{292}      

\newcommand{\fpgini}{0.79}          
\newcommand{\topfixn}{18}           
\newcommand{\mutextratasks}{74}     
\newcommand{\mcnemarp}{5.3\times10^{-23}} 
\newcommand{\staticzeromedian}{14}  
\newcommand{\escstat}{14.95}        
\newcommand{\escnat}{14.92}         

\newcommand{\contsteps}{800}        
\newcommand{\contseeds}{3}          
\newcommand{\contruns}{12}          
\newcommand{\contBslope}{-0.009}    
\newcommand{\contBslopeci}{[-0.33,\,+0.31]} 
\newcommand{\contCslope}{-0.008}    
\newcommand{\contCslopeci}{[-0.77,\,+0.75]} 
\newcommand{\contBhh}{-0.20}        
\newcommand{\contBhhci}{[-1.77,\,+1.38]} 
\newcommand{\contChh}{-0.37}        
\newcommand{\contChhci}{[-1.22,\,+0.48]} 
\newcommand{\contBmass}{-0.12}      
\newcommand{\contCmass}{-0.15}      
\newcommand{\contBtp}{+0.75}        
\newcommand{\contBtpci}{[+0.41,\,+1.09]} 
\newcommand{\contCtp}{+1.26}        
\newcommand{\contCtpci}{[+0.84,\,+1.68]} 
\newcommand{\contMDEB}{+0.43}       
\newcommand{\contMDEC}{+1.0}        
\newcommand{\contCmiss}{63}         
\newcommand{\contCrate}{+0.155}     
\newcommand{\contCratecifew}{[+0.061,\,+0.250]} 
\newcommand{\contCrateciols}{[-0.006,\,+0.317]} 

\newcommand{\figrhoA}{0.803}     
\newcommand{\figrhoB}{0.813}     
\newcommand{\figrhoC}{0.700}     
\newcommand{\figrhoAcond}{0.920} 
\newcommand{\figrhoBcond}{0.870} 
\newcommand{\figrhoCcond}{0.787} 
\newcommand{\fignAcond}{73}      
\newcommand{\fignBcond}{68}      
\newcommand{\fignCcond}{56}      
\newcommand{\fignB}{200}         
\newcommand{\fignC}{164}         
\newcommand{\figzeroA}{143}      
\newcommand{\figzeroB}{132}      
\newcommand{\figzeroC}{108}      
\newcommand{\figpceil}{10^{-12}} 
\newcommand{\outtaskA}{Mbpp/638} 
\newcommand{\outtaskB}{Mbpp/160} 
\newcommand{\outtaskC}{Mbpp/116} 
\newcommand{\outubfour}{0.60}    
\newcommand{\outubone}{0.98}     
\newcommand{\outtaskband}{Mbpp/459} 

\newcommand{\vfiveCagg}{-0.250}     
\newcommand{\vfiveCaggci}{[-0.490,\,-0.010]} 
\newcommand{\vfiveCearly}{-1.33}    
\newcommand{\vfiveCearlyci}{[-2.36,\,-0.30]} 
\newcommand{\vfivebaseearly}{+5.49} 
\newcommand{\vfivedilution}{-1.48}  
\newcommand{\vfiveCgh}{-0.16}       
\newcommand{\vfiveCghci}{[-0.61,\,+0.28]} 
\newcommand{\vfiveoldg}{-0.43}      
\newcommand{\vfiveanchor}{-0.249}   
\newcommand{\vfiverepl}{-0.2498}    
\newcommand{\vfiveBflat}{-0.09}     
\newcommand{\vfiveBflatci}{[-0.22,\,+0.04]} 
\newcommand{\vfivenovelB}{24}       
\newcommand{\vfivenovelC}{13}       
\newcommand{\vfivenovelBshare}{3.8} 
\newcommand{\vfivenovelCshare}{1.3} 

\newcommand{\probemaxdiff}{0.006}   
\newcommand{\probeauclo}{0.52}      
\newcommand{\probeauchi}{0.54}      

\newcommand{\basetestsavg}{3.1}     
\newcommand{\extratestsavg}{105.4}  
\newcommand{\disjointmbpp}{313}     
\newcommand{\mbpptotal}{378}        
\newcommand{\disjointhe}{71}        
\newcommand{\hetotal}{164}          
\newcommand{\relabelruns}{21}       
\newcommand{\relabelrewardn}{987}   
\newcommand{\relabelrewardN}{70{,}431} 
\newcommand{\relabelrewardpct}{1.40}   
\newcommand{\relabelmainn}{494}     
\newcommand{\relabelmainN}{39{,}584}   
\newcommand{\relabelmainpct}{1.25}  
\newcommand{\relabelsig}{853}       
\newcommand{\relabeltasks}{25}      
\newcommand{\probetbBprim}{+0.0003} 
\newcommand{\probetbCprim}{-0.0021} 
\newcommand{\probetbBrb}{+0.0002}   
\newcommand{\probetbBrtwo}{-0.0018} 
\newcommand{\probetbCrb}{-0.0008}   
\newcommand{\probetbCrtwo}{-0.0060} 
\newcommand{\probelmretB}{21.2}     
\newcommand{\probelmretC}{17.1}     
\newcommand{\probetbCrtwolm}{+0.0013} 
\newcommand{\bthreeshiftlo}{0.2}    
\newcommand{\bthreeshifthi}{0.9}    
\newcommand{\bthreearmdiff}{0.102}  

\newcommand{\fjgptB}{.959}   
\newcommand{\fjgptC}{.958}   
\newcommand{\fjdsB}{.902}    
\newcommand{\fjdsC}{.885}    
\newcommand{\fjlenB}{.672}   
\newcommand{\fjlenC}{.634}   

\newcommand{\maSelfDS}{.617}          
\newcommand{\maSelfDSci}{[.555,\,.679]}
\newcommand{\maSelfMM}{.639}          
\newcommand{\maSelfMMci}{[.590,\,.690]}
\newcommand{\maAnchDS}{.894}          
\newcommand{\maAnchDSci}{[.847,\,.936]}
\newcommand{\maAnchMM}{.786}          
\newcommand{\maAnchMMci}{[.710,\,.860]}
\newcommand{\maDiffDS}{-.276}         
\newcommand{\maDiffDSci}{[-.352,\,-.200]}
\newcommand{\maDiffMM}{-.148}         
\newcommand{\maDiffMMcitab}{[-.236,\,-.058]} 
\newcommand{\maDiffMMcilic}{[-.236,\,-.057]} 
\newcommand{\maBtwDS}{84.8}           
\newcommand{\maBtwMM}{77.7}           
\newcommand{\maRhoDS}{.605}           
\newcommand{\maRhoMM}{.631}           
\newcommand{\maBand}{.592--.639}      
\newcommand{\maDeltaDS}{-.025}        
\newcommand{\maDeltaDSci}{[-.090,\,+.040]}
\newcommand{\maDeltaMM}{-.048}        
\newcommand{\maDeltaMMci}{[-.109,\,+.012]}
\newcommand{\maIDS}{-.106}            
\newcommand{\maIDSci}{[-.178,\,-.035]}
\newcommand{\maIDSp}{.0032}
\newcommand{\maIMM}{+.021}            
\newcommand{\maIMMci}{[-.020,\,+.064]}
\newcommand{\maIMMp}{.32}

%% file: appendix_tables.tex

\begin{table}[!htbp]
\centering
\small
\caption{Checkpoint-400 leaky--hardened pass-rate gap under six 90\% CI
estimators (P1). The point estimate common to all six intervals is the
percentile gap mean $=0.197$~pt (\texttt{ckpts/400/gap\_mean\_pt}); the
one-sided 95\% upper bound equals the two-sided 90\% CI upper endpoint.
All values in percentage points.}
\label{tab:app-p1-estimators}
\begin{tabular}{lcc}
\toprule
Estimator & 90\% CI (pt) & One-sided 95\% upper (pt) \\
\midrule
Percentile & $[-0.420,\ 0.745]$ & $0.745$ \\
Basic & $[-0.351,\ 0.813]$ & $0.813$ \\
BCa & $[-0.437,\ 0.702]$ & $0.702$ \\
Hierarchical & $[-0.488,\ 0.873]$ & $0.873$ \\
Seed-$t$ & $[-0.606,\ 1.000]$ & $1.000$ \\
Sign-flip & $[-0.728,\ 1.070]$ & $1.070$ \\
\midrule
Primary gap mean (point) & \multicolumn{2}{c}{$0.197$~pt (percentile point est., ckpt400)} \\
\bottomrule
\end{tabular}
\end{table}

\begin{table}[!htbp]
\centering
\small
\caption{Held-out pass-rate gap (H$-$L, percentage points) by static-leakiness
stratum across training checkpoints (P3, exploratory). Positive = leaky arm
worse. Stratum labels carry their task count (e.g.\ \texttt{exposed47} $=47$
tasks); \texttt{null16} is the negative-control stratum. Source column order is
checkpoints 100/200/300/400. The apparent deepening across checkpoints in the
exposed rows is a seed-pooled descriptive aggregate; per seed it is monotone in
one of five seeds (frozen rider; see \S4.4).}
\label{tab:app-p3-strata}
\begin{tabular}{lrrrr}
\toprule
Stratum & ckpt 100 & ckpt 200 & ckpt 300 & ckpt 400 \\
\midrule
\texttt{exposed47} & 0.11 & 0.74 & 0.90 & 1.60 \\
\texttt{clean81} & -0.25 & 0.00 & 0.09 & -0.03 \\
\texttt{clean65} & -0.23 & 0.04 & 0.19 & 0.08 \\
\texttt{midband27} & 0.00 & 0.93 & 1.48 & 2.78 \\
\texttt{highleak20} & 0.25 & 0.50 & 0.12 & 0.00 \\
\texttt{floor30} & 0.08 & 0.75 & 1.42 & 2.58 \\
\texttt{null16} & -0.31 & -0.16 & -0.31 & -0.47 \\
\bottomrule
\end{tabular}
\end{table}

\begin{center}
\footnotesize
\begin{longtable}{p{0.5\textwidth}rrrr}
\caption{Spec-search grid: all 106 valid stratifying specifications at ckpt400,
each defining an exposed stratum $S$ vs.\ its complement. Columns: spec
definition; stratum size $n_S$; leaky--hardened gap (H$-$L, pt); seed-$t$
statistic; and rank of $t$ among the 106 specs (competition ranking, 1 = largest
$t$). Share of specs with positive $t$: 98.11\%. Family-adjusted one-sided
$p$ (primary seed-$t$ statistic) under the three nulls: N1a family-$t$ 0.0913, best-$t$ 0.0357; N1b family-$t$ 0.0938, best-$t$ 0.0625; N2 family-$t$ 0.3933, best-$t$ 0.1447. Rows marked
$^{\dagger}$ are the three pre-registered / historical anchor strata
(\texttt{exposed47}, \texttt{midband27}, \texttt{floor30}).}
\label{tab:app-specsearch} \\
\toprule
Spec (exposed-stratum definition) & $n_S$ & gap (pt) & $t$ & rank \\
\midrule
\endfirsthead
\multicolumn{5}{c}{\tablename\ \thetable\ -- continued} \\
\toprule
Spec (exposed-stratum definition) & $n_S$ & gap (pt) & $t$ & rank \\
\midrule
\endhead
\midrule \multicolumn{5}{r}{\emph{continued on next page}} \\
\endfoot
\bottomrule
\endlastfoot
\texttt{L\textgreater{}0.0 \textbar{} None-\textgreater{}complement \textbar{} floor=none}$^{\dagger}$ & 47 & 1.60 & 2.77 & 43 \\
\texttt{L\textgreater{}0.0 \textbar{} None-\textgreater{}complement \textbar{} floor=gt0}$^{\dagger}$ & 30 & 2.58 & 3.08 & 27 \\
\texttt{L\textgreater{}0.0 \textbar{} None-\textgreater{}complement \textbar{} floor=ge025} & 26 & 2.60 & 2.75 & 47 \\
\texttt{L\textgreater{}0.0 \textbar{} None-\textgreater{}excluded \textbar{} floor=none} & 47 & 1.60 & 2.77 & 43 \\
\texttt{L\textgreater{}0.0 \textbar{} None-\textgreater{}excluded \textbar{} floor=gt0} & 30 & 2.58 & 3.08 & 27 \\
\texttt{L\textgreater{}0.0 \textbar{} None-\textgreater{}excluded \textbar{} floor=ge025} & 26 & 2.60 & 2.75 & 47 \\
\texttt{L\textgreater{}0.1 \textbar{} None-\textgreater{}complement \textbar{} floor=none} & 39 & 1.54 & 3.45 & 15 \\
\texttt{L\textgreater{}0.1 \textbar{} None-\textgreater{}complement \textbar{} floor=gt0} & 22 & 2.84 & 3.95 & 9 \\
\texttt{L\textgreater{}0.1 \textbar{} None-\textgreater{}complement \textbar{} floor=ge025} & 18 & 2.92 & 3.38 & 19 \\
\texttt{L\textgreater{}0.1 \textbar{} None-\textgreater{}excluded \textbar{} floor=none} & 39 & 1.54 & 3.45 & 15 \\
\texttt{L\textgreater{}0.1 \textbar{} None-\textgreater{}excluded \textbar{} floor=gt0} & 22 & 2.84 & 3.95 & 9 \\
\texttt{L\textgreater{}0.1 \textbar{} None-\textgreater{}excluded \textbar{} floor=ge025} & 18 & 2.92 & 3.38 & 19 \\
\texttt{L\textgreater{}0.2 \textbar{} None-\textgreater{}complement \textbar{} floor=none} & 29 & 1.21 & 1.34 & 87 \\
\texttt{L\textgreater{}0.2 \textbar{} None-\textgreater{}complement \textbar{} floor=gt0} & 12 & 3.13 & 1.70 & 81 \\
\texttt{L\textgreater{}0.2 \textbar{} None-\textgreater{}complement \textbar{} floor=ge025} & 10 & 3.50 & 1.91 & 73 \\
\texttt{L\textgreater{}0.2 \textbar{} None-\textgreater{}excluded \textbar{} floor=none} & 29 & 1.21 & 1.34 & 87 \\
\texttt{L\textgreater{}0.2 \textbar{} None-\textgreater{}excluded \textbar{} floor=gt0} & 12 & 3.13 & 1.70 & 81 \\
\texttt{L\textgreater{}0.2 \textbar{} None-\textgreater{}excluded \textbar{} floor=ge025} & 10 & 3.50 & 1.91 & 73 \\
\texttt{L\textgreater{}0.3 \textbar{} None-\textgreater{}complement \textbar{} floor=none} & 26 & 0.96 & 1.07 & 99 \\
\texttt{L\textgreater{}0.3 \textbar{} None-\textgreater{}excluded \textbar{} floor=none} & 26 & 0.96 & 1.07 & 99 \\
\texttt{L\textgreater{}0.4 \textbar{} None-\textgreater{}complement \textbar{} floor=none} & 23 & 0.98 & 1.15 & 97 \\
\texttt{L\textgreater{}0.4 \textbar{} None-\textgreater{}excluded \textbar{} floor=none} & 23 & 0.98 & 1.15 & 97 \\
\texttt{L\textgreater{}0.5 \textbar{} None-\textgreater{}complement \textbar{} floor=none} & 20 & 0.00 & 0.00 & 105 \\
\texttt{L\textgreater{}0.5 \textbar{} None-\textgreater{}excluded \textbar{} floor=none} & 20 & 0.00 & 0.00 & 105 \\
\texttt{0.0\textless{}L\textless{}=0.3 \textbar{} None-\textgreater{}complement \textbar{} floor=none} & 21 & 2.38 & 2.53 & 57 \\
\texttt{0.0\textless{}L\textless{}=0.3 \textbar{} None-\textgreater{}complement \textbar{} floor=gt0} & 21 & 2.38 & 2.53 & 57 \\
\texttt{0.0\textless{}L\textless{}=0.3 \textbar{} None-\textgreater{}complement \textbar{} floor=ge025} & 19 & 2.24 & 2.31 & 65 \\
\texttt{0.0\textless{}L\textless{}=0.3 \textbar{} None-\textgreater{}excluded \textbar{} floor=none} & 21 & 2.38 & 2.53 & 57 \\
\texttt{0.0\textless{}L\textless{}=0.3 \textbar{} None-\textgreater{}excluded \textbar{} floor=gt0} & 21 & 2.38 & 2.53 & 57 \\
\texttt{0.0\textless{}L\textless{}=0.3 \textbar{} None-\textgreater{}excluded \textbar{} floor=ge025} & 19 & 2.24 & 2.31 & 65 \\
\texttt{0.0\textless{}L\textless{}=0.4 \textbar{} None-\textgreater{}complement \textbar{} floor=none} & 24 & 2.19 & 2.87 & 33 \\
\texttt{0.0\textless{}L\textless{}=0.4 \textbar{} None-\textgreater{}complement \textbar{} floor=gt0} & 24 & 2.19 & 2.87 & 33 \\
\texttt{0.0\textless{}L\textless{}=0.4 \textbar{} None-\textgreater{}complement \textbar{} floor=ge025} & 22 & 2.05 & 2.57 & 55 \\
\texttt{0.0\textless{}L\textless{}=0.4 \textbar{} None-\textgreater{}excluded \textbar{} floor=none} & 24 & 2.19 & 2.87 & 33 \\
\texttt{0.0\textless{}L\textless{}=0.4 \textbar{} None-\textgreater{}excluded \textbar{} floor=gt0} & 24 & 2.19 & 2.87 & 33 \\
\texttt{0.0\textless{}L\textless{}=0.4 \textbar{} None-\textgreater{}excluded \textbar{} floor=ge025} & 22 & 2.05 & 2.57 & 55 \\
\texttt{0.0\textless{}L\textless{}=0.5 \textbar{} None-\textgreater{}complement \textbar{} floor=none}$^{\dagger}$ & 27 & 2.78 & 3.59 & 11 \\
\texttt{0.0\textless{}L\textless{}=0.5 \textbar{} None-\textgreater{}complement \textbar{} floor=gt0} & 27 & 2.78 & 3.59 & 11 \\
\texttt{0.0\textless{}L\textless{}=0.5 \textbar{} None-\textgreater{}complement \textbar{} floor=ge025} & 25 & 2.70 & 3.09 & 25 \\
\texttt{0.0\textless{}L\textless{}=0.5 \textbar{} None-\textgreater{}excluded \textbar{} floor=none} & 27 & 2.78 & 3.59 & 11 \\
\texttt{0.0\textless{}L\textless{}=0.5 \textbar{} None-\textgreater{}excluded \textbar{} floor=gt0} & 27 & 2.78 & 3.59 & 11 \\
\texttt{0.0\textless{}L\textless{}=0.5 \textbar{} None-\textgreater{}excluded \textbar{} floor=ge025} & 25 & 2.70 & 3.09 & 25 \\
\texttt{0.0\textless{}L\textless{}=0.75 \textbar{} None-\textgreater{}complement \textbar{} floor=none} & 34 & 2.21 & 2.77 & 41 \\
\texttt{0.0\textless{}L\textless{}=0.75 \textbar{} None-\textgreater{}complement \textbar{} floor=gt0} & 29 & 2.67 & 3.08 & 27 \\
\texttt{0.0\textless{}L\textless{}=0.75 \textbar{} None-\textgreater{}complement \textbar{} floor=ge025} & 26 & 2.60 & 2.75 & 47 \\
\texttt{0.0\textless{}L\textless{}=0.75 \textbar{} None-\textgreater{}excluded \textbar{} floor=none} & 34 & 2.21 & 2.77 & 41 \\
\texttt{0.0\textless{}L\textless{}=0.75 \textbar{} None-\textgreater{}excluded \textbar{} floor=gt0} & 29 & 2.67 & 3.08 & 27 \\
\texttt{0.0\textless{}L\textless{}=0.75 \textbar{} None-\textgreater{}excluded \textbar{} floor=ge025} & 26 & 2.60 & 2.75 & 47 \\
\texttt{0.1\textless{}L\textless{}=0.3 \textbar{} None-\textgreater{}complement \textbar{} floor=none} & 13 & 2.69 & 2.26 & 67 \\
\texttt{0.1\textless{}L\textless{}=0.3 \textbar{} None-\textgreater{}complement \textbar{} floor=gt0} & 13 & 2.69 & 2.26 & 67 \\
\texttt{0.1\textless{}L\textless{}=0.3 \textbar{} None-\textgreater{}complement \textbar{} floor=ge025} & 11 & 2.50 & 2.40 & 63 \\
\texttt{0.1\textless{}L\textless{}=0.3 \textbar{} None-\textgreater{}excluded \textbar{} floor=none} & 13 & 2.69 & 2.26 & 67 \\
\texttt{0.1\textless{}L\textless{}=0.3 \textbar{} None-\textgreater{}excluded \textbar{} floor=gt0} & 13 & 2.69 & 2.26 & 67 \\
\texttt{0.1\textless{}L\textless{}=0.3 \textbar{} None-\textgreater{}excluded \textbar{} floor=ge025} & 11 & 2.50 & 2.40 & 63 \\
\texttt{0.1\textless{}L\textless{}=0.4 \textbar{} None-\textgreater{}complement \textbar{} floor=none} & 16 & 2.34 & 2.86 & 37 \\
\texttt{0.1\textless{}L\textless{}=0.4 \textbar{} None-\textgreater{}complement \textbar{} floor=gt0} & 16 & 2.34 & 2.86 & 37 \\
\texttt{0.1\textless{}L\textless{}=0.4 \textbar{} None-\textgreater{}complement \textbar{} floor=ge025} & 14 & 2.14 & 3.21 & 23 \\
\texttt{0.1\textless{}L\textless{}=0.4 \textbar{} None-\textgreater{}excluded \textbar{} floor=none} & 16 & 2.34 & 2.86 & 37 \\
\texttt{0.1\textless{}L\textless{}=0.4 \textbar{} None-\textgreater{}excluded \textbar{} floor=gt0} & 16 & 2.34 & 2.86 & 37 \\
\texttt{0.1\textless{}L\textless{}=0.4 \textbar{} None-\textgreater{}excluded \textbar{} floor=ge025} & 14 & 2.14 & 3.21 & 23 \\
\texttt{0.1\textless{}L\textless{}=0.5 \textbar{} None-\textgreater{}complement \textbar{} floor=none} & 19 & 3.16 & 4.71 & 1 \\
\texttt{0.1\textless{}L\textless{}=0.5 \textbar{} None-\textgreater{}complement \textbar{} floor=gt0} & 19 & 3.16 & 4.71 & 1 \\
\texttt{0.1\textless{}L\textless{}=0.5 \textbar{} None-\textgreater{}complement \textbar{} floor=ge025} & 17 & 3.09 & 4.33 & 5 \\
\texttt{0.1\textless{}L\textless{}=0.5 \textbar{} None-\textgreater{}excluded \textbar{} floor=none} & 19 & 3.16 & 4.71 & 1 \\
\texttt{0.1\textless{}L\textless{}=0.5 \textbar{} None-\textgreater{}excluded \textbar{} floor=gt0} & 19 & 3.16 & 4.71 & 1 \\
\texttt{0.1\textless{}L\textless{}=0.5 \textbar{} None-\textgreater{}excluded \textbar{} floor=ge025} & 17 & 3.09 & 4.33 & 5 \\
\texttt{0.1\textless{}L\textless{}=0.75 \textbar{} None-\textgreater{}complement \textbar{} floor=none} & 26 & 2.31 & 3.45 & 15 \\
\texttt{0.1\textless{}L\textless{}=0.75 \textbar{} None-\textgreater{}complement \textbar{} floor=gt0} & 21 & 2.98 & 3.95 & 7 \\
\texttt{0.1\textless{}L\textless{}=0.75 \textbar{} None-\textgreater{}complement \textbar{} floor=ge025} & 18 & 2.92 & 3.38 & 19 \\
\texttt{0.1\textless{}L\textless{}=0.75 \textbar{} None-\textgreater{}excluded \textbar{} floor=none} & 26 & 2.31 & 3.45 & 15 \\
\texttt{0.1\textless{}L\textless{}=0.75 \textbar{} None-\textgreater{}excluded \textbar{} floor=gt0} & 21 & 2.98 & 3.95 & 7 \\
\texttt{0.1\textless{}L\textless{}=0.75 \textbar{} None-\textgreater{}excluded \textbar{} floor=ge025} & 18 & 2.92 & 3.38 & 19 \\
\texttt{0.2\textless{}L\textless{}=0.75 \textbar{} None-\textgreater{}complement \textbar{} floor=none} & 16 & 2.19 & 1.34 & 91 \\
\texttt{0.2\textless{}L\textless{}=0.75 \textbar{} None-\textgreater{}complement \textbar{} floor=gt0} & 11 & 3.41 & 1.70 & 79 \\
\texttt{0.2\textless{}L\textless{}=0.75 \textbar{} None-\textgreater{}complement \textbar{} floor=ge025} & 10 & 3.50 & 1.91 & 73 \\
\texttt{0.2\textless{}L\textless{}=0.75 \textbar{} None-\textgreater{}excluded \textbar{} floor=none} & 16 & 2.19 & 1.34 & 91 \\
\texttt{0.2\textless{}L\textless{}=0.75 \textbar{} None-\textgreater{}excluded \textbar{} floor=gt0} & 11 & 3.41 & 1.70 & 79 \\
\texttt{0.2\textless{}L\textless{}=0.75 \textbar{} None-\textgreater{}excluded \textbar{} floor=ge025} & 10 & 3.50 & 1.91 & 73 \\
\texttt{0.2\textless{}L\textless{}=1.0 \textbar{} None-\textgreater{}complement \textbar{} floor=none} & 29 & 1.21 & 1.34 & 87 \\
\texttt{0.2\textless{}L\textless{}=1.0 \textbar{} None-\textgreater{}complement \textbar{} floor=gt0} & 12 & 3.13 & 1.70 & 81 \\
\texttt{0.2\textless{}L\textless{}=1.0 \textbar{} None-\textgreater{}complement \textbar{} floor=ge025} & 10 & 3.50 & 1.91 & 73 \\
\texttt{0.2\textless{}L\textless{}=1.0 \textbar{} None-\textgreater{}excluded \textbar{} floor=none} & 29 & 1.21 & 1.34 & 87 \\
\texttt{0.2\textless{}L\textless{}=1.0 \textbar{} None-\textgreater{}excluded \textbar{} floor=gt0} & 12 & 3.13 & 1.70 & 81 \\
\texttt{0.2\textless{}L\textless{}=1.0 \textbar{} None-\textgreater{}excluded \textbar{} floor=ge025} & 10 & 3.50 & 1.91 & 73 \\
\texttt{0.3\textless{}L\textless{}=0.75 \textbar{} None-\textgreater{}complement \textbar{} floor=none} & 13 & 1.92 & 1.07 & 99 \\
\texttt{0.3\textless{}L\textless{}=0.75 \textbar{} None-\textgreater{}excluded \textbar{} floor=none} & 13 & 1.92 & 1.07 & 99 \\
\texttt{0.3\textless{}L\textless{}=1.0 \textbar{} None-\textgreater{}complement \textbar{} floor=none} & 26 & 0.96 & 1.07 & 99 \\
\texttt{0.3\textless{}L\textless{}=1.0 \textbar{} None-\textgreater{}excluded \textbar{} floor=none} & 26 & 0.96 & 1.07 & 99 \\
\texttt{FP\textgreater{}=1 \textbar{} None-\textgreater{}complement \textbar{} floor=none} & 47 & 1.60 & 2.77 & 43 \\
\texttt{FP\textgreater{}=1 \textbar{} None-\textgreater{}complement \textbar{} floor=gt0} & 30 & 2.58 & 3.08 & 27 \\
\texttt{FP\textgreater{}=1 \textbar{} None-\textgreater{}complement \textbar{} floor=ge025} & 26 & 2.60 & 2.75 & 47 \\
\texttt{FP\textgreater{}=1 \textbar{} None-\textgreater{}excluded \textbar{} floor=none} & 47 & 1.60 & 2.77 & 43 \\
\texttt{FP\textgreater{}=1 \textbar{} None-\textgreater{}excluded \textbar{} floor=gt0} & 30 & 2.58 & 3.08 & 27 \\
\texttt{FP\textgreater{}=1 \textbar{} None-\textgreater{}excluded \textbar{} floor=ge025} & 26 & 2.60 & 2.75 & 47 \\
\texttt{FP\textgreater{}=2 \textbar{} None-\textgreater{}complement \textbar{} floor=none} & 32 & 1.48 & 2.14 & 71 \\
\texttt{FP\textgreater{}=2 \textbar{} None-\textgreater{}complement \textbar{} floor=gt0} & 18 & 2.78 & 2.53 & 57 \\
\texttt{FP\textgreater{}=2 \textbar{} None-\textgreater{}complement \textbar{} floor=ge025} & 16 & 2.97 & 2.73 & 53 \\
\texttt{FP\textgreater{}=2 \textbar{} None-\textgreater{}excluded \textbar{} floor=none} & 32 & 1.48 & 2.14 & 71 \\
\texttt{FP\textgreater{}=2 \textbar{} None-\textgreater{}excluded \textbar{} floor=gt0} & 18 & 2.78 & 2.53 & 57 \\
\texttt{FP\textgreater{}=2 \textbar{} None-\textgreater{}excluded \textbar{} floor=ge025} & 16 & 2.97 & 2.73 & 53 \\
\texttt{FP\textgreater{}=3 \textbar{} None-\textgreater{}complement \textbar{} floor=none} & 25 & 1.10 & 1.15 & 95 \\
\texttt{FP\textgreater{}=3 \textbar{} None-\textgreater{}complement \textbar{} floor=gt0} & 11 & 2.73 & 1.47 & 85 \\
\texttt{FP\textgreater{}=3 \textbar{} None-\textgreater{}excluded \textbar{} floor=none} & 25 & 1.10 & 1.15 & 95 \\
\texttt{FP\textgreater{}=3 \textbar{} None-\textgreater{}excluded \textbar{} floor=gt0} & 11 & 2.73 & 1.47 & 85 \\
\texttt{FP\textgreater{}=5 \textbar{} None-\textgreater{}complement \textbar{} floor=none} & 16 & 1.41 & 1.18 & 93 \\
\texttt{FP\textgreater{}=5 \textbar{} None-\textgreater{}excluded \textbar{} floor=none} & 16 & 1.41 & 1.18 & 93 \\
\end{longtable}
\end{center}

\begin{table}[!htbp]
\centering
\small
\caption{Per-seed verified-wrong (vw) count over the leaky false-positive pool,
all three model families. \texttt{vw/pool} is the count of records with
\texttt{verified\_wrong}$=$true over the leaky FP-pool records at that seed;
share is their ratio. Family-A seeds are recovered by re-joining the leaky FP
rollouts (\texttt{base}$=$True, \texttt{plus}$=$False) to the per-code v5
verdict table (frozen audit method \texttt{seed\_join.py}); this reproduces the
pooled total exactly with \texttt{join\_missing}$=0$ and
\texttt{verdict\_conflict}$=0$ (no seed-ambiguous records). Pooled anchors
1342/2821 (A), 976/2151 (B), 1366/2176 (C) all reproduce.}
\label{tab:app-vw-perseed}
\begin{tabular}{llrr}
\toprule
Family & Seed & vw/pool & share (\%) \\
\midrule
A (Qwen2.5, train grid) & 0 & 265/562 & 47.15 \\
 & 1 & 257/563 & 45.65 \\
 & 2 & 270/568 & 47.54 \\
 & 3 & 262/548 & 47.81 \\
 & 4 & 288/580 & 49.66 \\
\textbf{A (Qwen2.5, train grid)} & \textbf{pooled} & \textbf{1342/2821} & \textbf{47.57} \\
\midrule
B (DeepSeek-Coder-1.3B) & 0 & 188/422 & 44.55 \\
 & 1 & 213/449 & 47.44 \\
 & 2 & 202/427 & 47.31 \\
 & 3 & 191/419 & 45.58 \\
 & 4 & 182/434 & 41.94 \\
\textbf{B (DeepSeek-Coder-1.3B)} & \textbf{pooled} & \textbf{976/2151} & \textbf{45.37} \\
\midrule
C (Llama-3.2-1B) & 0 & 283/448 & 63.17 \\
 & 1 & 248/403 & 61.54 \\
 & 2 & 270/425 & 63.53 \\
 & 3 & 283/458 & 61.79 \\
 & 4 & 282/442 & 63.80 \\
\textbf{C (Llama-3.2-1B)} & \textbf{pooled} & \textbf{1366/2176} & \textbf{62.78} \\
\bottomrule
\end{tabular}
\end{table}

\begin{table}[!htbp]
\centering
\small
\caption{Kill-test A, per fingerprint channel (family-mean denominators). For
each channel: base-model match rate on the leaky FP subset (\texttt{base==True,
plus==False}); base match rate over all samples; leaky-trained match rate; and
the family-mean kill verdict (base FP-subset $\ge$ trained $\Rightarrow$
KILLED). Family-mean gives KILLED 9/11; the pooled-across-families denominator
gives 7/11 and the min-denominator-guarded ($n_{\mathrm{fp}}\ge5$) denominator
gives 10/11.}
\label{tab:app-killtestA}
\begin{tabular}{lcccl}
\toprule
Channel & base FP-subset (\%) & base all (\%) & trained (\%) & verdict \\
\midrule
\texttt{Mbpp/771} & 100.00 & 60.42 & 100.00 & KILLED \\
\texttt{Mbpp/99} & 63.70 & 29.17 & 61.03 & KILLED \\
\texttt{Mbpp/576} & 100.00 & 68.75 & 97.03 & KILLED \\
\texttt{Mbpp/435} & 77.78 & 50.00 & 89.69 & SURVIVES \\
\texttt{Mbpp/267} & 97.62 & 72.92 & 93.41 & KILLED \\
\texttt{Mbpp/287} & 100.00 & 47.92 & 100.00 & KILLED \\
\texttt{Mbpp/790} & 100.00 & 39.58 & 97.08 & KILLED \\
\texttt{Mbpp/639} & 50.00 & 37.50 & 94.50 & SURVIVES \\
\texttt{Mbpp/806} & 100.00 & 50.00 & 98.96 & KILLED \\
\texttt{Mbpp/781} & 100.00 & 47.92 & 100.00 & KILLED \\
\texttt{Mbpp/786} & 68.69 & 47.92 & 68.09 & KILLED \\
\bottomrule
\end{tabular}
\end{table}

\begin{table}[!htbp]
\centering
\footnotesize
\caption{R3 failure-mode symmetry (family B). $\delta = S_{\mathrm{leaky}} -
S_{\mathrm{hard}}$ is the difference in eval-failure-profile divergence from the
train profile (dimensionless, JSD-based); positive $\delta$ = leaky-arm eval
failures diverge more. Primary reading (per-test, non-artifact classes) on the
exposed stratum with its clean-stratum control, and the primary-tag sensitivity
reading. Bootstrap 90\% CI ($B{=}4000$), sign-flip one-sided $p$ (floor
$0.03125$), and the per-seed sign pattern (seeds 0--4).}
\label{tab:app-r3}
\setlength{\tabcolsep}{3.5pt}
\begin{tabular}{lrrrc}
\toprule
Reading / stratum & $\delta$ & bootstrap 90\% CI & sign-flip $p$ & per-seed signs \\
\midrule
Primary (per-test non-artifact), exposed & $0.00003$ & $[-0.0104,\ 0.0147]$ & $0.5625$ & $-$/$+$/$-$/$-$/$+$ \\
\quad primary reading, clean stratum & $0.00051$ & $[-0.0104,\ 0.0125]$ & $0.2812$ & $-$/$+$/$+$/$+$/$+$ \\
Sensitivity (primary-tag), exposed & $0.01704$ & $[-0.0009,\ 0.0349]$ & $0.0312$ & $+$/$+$/$+$/$+$/$+$ \\
\bottomrule
\end{tabular}
\end{table}

\begin{table}[!htbp]
\centering
\scriptsize
\setlength{\tabcolsep}{2.5pt}
\caption{Preregistered 800-step continuation (fresh runs, families B and C,
\contseeds\ seeds per arm): per-run slopes in pp per 100 steps, OLS with CI90
per seed; family-aggregate rows are seed-level $t$ intervals (df$=2$), the
decision level. Columns: FP-only-group share slope over the full window;
second-half minus first-half difference (full second-half intervals are in
the released metrics file); rewarded-FP mass slope (secondary criterion);
genuine-pass rate slope. No FP-only share
slope excludes zero on the positive side in any of the \contruns\ runs, while
genuine-pass rates rise throughout. Power is asymmetric by design: family B's
MDE at 80\% power is \contMDEB\,pp per 100 steps (a powered refutation);
family C's is \contMDEC\ (an underpowered non-detection). The continuation
reuses the same 250 training tasks (seed axis refreshed, task axis untested).}
\label{tab:app-cont800}
\begin{tabular}{lcccc}
\toprule
Run & FP-only share, full & $\Delta$(h2$-$h1) & rewarded-FP mass, full & TP rate, full \\
\midrule
B s0 & +0.191 [-0.483,+0.865] & +0.017 & -0.220 [-0.548,+0.108] & +0.952 [+0.328,+1.575] \\
B s1 & -0.036 [-0.713,+0.641] & +0.611 & +0.151 [-0.182,+0.484] & +0.552 [-0.068,+1.172] \\
B s2 & -0.182 [-0.839,+0.475] & -1.221 & -0.297 [-0.628,+0.034] & +0.742 [+0.127,+1.357] \\
\textbf{B (family agg.)} & \textbf{-0.009 [-0.326,+0.307]} & \textbf{-0.197 [-1.773,+1.378]} & \textbf{-0.122 [-0.526,+0.281]} & \textbf{+0.749 [+0.411,+1.085]} \\
\midrule
C s0 & -0.077 [-0.799,+0.645] & -0.489 & -0.308 [-0.653,+0.037] & +1.548 [+0.877,+2.220] \\
C s1 & +0.472 [-0.258,+1.202] & +0.185 & +0.110 [-0.233,+0.454] & +1.107 [+0.430,+1.785] \\
C s2 & -0.421 [-1.122,+0.281] & -0.799 & -0.243 [-0.582,+0.096] & +1.129 [+0.447,+1.810] \\
\textbf{C (family agg.)} & \textbf{-0.008 [-0.768,+0.751]} & \textbf{-0.367 [-1.216,+0.481]} & \textbf{-0.147 [-0.526,+0.233]} & \textbf{+1.262 [+0.842,+1.681]} \\
\bottomrule
\end{tabular}
\end{table}